%% file: arxiv_main.tex
\documentclass[]{fairmeta}
\usepackage{xcolor}
\usepackage[normalem]{ulem} \definecolor{deletionblue}{RGB}{125,145,175}

\usepackage{comment}
\usepackage{amsmath,amsfonts,bm,bbm}
\usepackage{amsthm}

\usepackage{multirow}
\usepackage{colortbl}
\usepackage{tabularx}
\usepackage{algorithm}
\usepackage[noend]{algpseudocode}
\newcolumntype{Y}{>{\centering\arraybackslash}X}
\usepackage{fontawesome}
\usepackage[bottom]{footmisc}
\usepackage{enumerate}
\usepackage{hyperref}
\usepackage{enumitem}
\usepackage{soul}

\usepackage{tcolorbox}
\definecolor{MistyRose}{rgb}{0.99, 0.91, 0.95}
\definecolor{myyellow}{rgb}{1,0.96,0.56}
\definecolor{improvegreen}{rgb}{0.18,0.49,0.20}
\definecolor{regressred}{rgb}{0.70,0.15,0.12}

\def\ouralg{\texttt{LIGE-GR}}
\DeclareMathOperator*{\argmax}{arg\,max}

\newcommand{\prodA}{Instagram Reels}
\newcommand{\prodB}{Facebook Video}
\newcommand{\palette}{\texttt{Palette}}
\newcommand{\impchg}[1]{\textcolor{improvegreen}{\textbf{#1}}}
\newcommand{\regchg}[1]{\textcolor{regressred}{\textbf{#1}}}

\title{$\ouralg$: A Smooth Leap from Ranking to Generative Recommendation in the LLM Era}

\author{Venkat Srinivas$^\star$}
\author{Chenzhang He$^\star$}
\author{Sam Woodmansee$^\star$}
\author{Shawn Lian$^\star$}
\author{Wenjie Hu$^\star$}
\author{Renjie Jiang$^\star$}
\author{Ziheng Huang}
\author{Xinyuan Zhang}
\author{Zhihao Zheng}
\author{Zhuoran Yu}
\author{Rui Li}
\author{Lei Yuan}
\author{Ziwei Li}
\author{Jimmy Jia}
\author{Mert Terzihan} 
\author{Ekrem Kocaguneli}
\author{Yiming Liao}
\author{Zhichen Zhao}
\author{Yue Yin}
\author{Yue Weng}
\author{Wanli Ma}
\author{Xufeng Cai}
\author{Weimiao Wu}
\author{Yezhou Huang}
\author{Du Zhang}
\author{Yukun Ding}
\author{Aaron Johnston}
\author{Yueming Wang}
\author{Zhaojie Gong}
\author{Yuting Zhang}
\author{Serena Li}
\author{Adithya Ganesh}
\author{Boying Liu}
\author{Haichuan Yang}
\author{Xialu Li}
\author{Matt Ma}
\author{Qunshu Zhang}
\author{Andrew Ton}
\author{Nathan Berrebbi}
\author{Neel Pawar}
\author{John Joshua Miller}
\author{Jayant Subramanian}
\author{Praveen Rathinavelu}
\author{Cheng Huang}
\author{Aadhar Sachdeva}
\author{Josh Karns}
\author{Andres Aaron Gutierrez}
\author{Neil Agarwal}
\author{Gustas Pladis}
\author{Vladimir Batygin}
\author{Gopal Ray}
\author{Aditya Priyadarshi}
\author{Shantanu Patil}
\author{Zhe Wang}
\author{Penny Pan}
\author{Yiping Han}
\author{Arun Singh}
\author{Guangdeng Liao}
\author{Bi Xue}
\author{Xinyao Hu}
\author{Yang Song}
\author{Yisong Song}
\author{Meihong Wang}
\author{Haotian Wu}
\author{Deepak Agarwal}
\author{Ji Liu}

\affiliation{Meta Platforms, Inc.,\ Menlo Park, CA, USA}

\abstract{
The remarkable success of large language models (LLMs) has provided important inspiration for the next generation of recommender systems. Structurally, recommendation and language generation share a similarity: both aim to produce an ordered sequence that optimizes the user's experience. However, how to precisely absorb the essence of the LLM paradigm into mature industrial recommender systems remains an open problem.

There are two challenges. First, it is unclear how to incorporate the LLM paradigm — sequence-level generation and optimization — into recommendation. Second, real-world recommender systems are mature systems that have been iteratively customized for years around specific products, business constraints, serving infrastructure, and organizational ownership. Replacing such systems wholesale is often technically risky and organizationally disruptive. 

In this paper, we propose $\ouralg$, a listwise generation and evaluation recommendation framework that upgrades from a traditional ranking system (itemwise recommendation) toward a generative recommendation paradigm. Instead of rebuilding the entire recommendation stack from scratch, $\ouralg$ generalizes the existing pointwise recommendation system into a listwise generation system. This allows mature recommender systems to benefit from listwise optimization while preserving compatibility with existing models, value functions, and serving infrastructure. 

We validate $\ouralg$ in short-video recommendation on \prodA{} and \prodB{}. On these recommendation surfaces, $\ouralg$ improves time spent by $1.14\%$ on Instagram Reels and $0.72\%$ on Facebook Video, while requiring only modest additional inference resources.
}

\correspondence{Ji Liu at \email{madisonliu@meta.com}; $^\star$ equal contribution.}

\begin{document}

\maketitle

\section{Introduction}\label{sec:intro}

Modern recommender systems return an ordered list of content items for each user request, which are then sequentially exposed to the user. This formulation spans short-video and feed recommendation.
Despite their sophistication, many industrial recommender systems are still fundamentally built around itemwise optimization. A ranking model predicts user engagement signals for each candidate item independently, and a prespecified value model (function) maps these predicted signals into a scalar score. Items are then sorted by their individual scores to form a recommendation list. Additional product constraints, such as diversity or integrity adjustments, are usually introduced through heuristic or rule-based score modifications. This is the general framework that state-of-the-art products employ today. Although this paradigm has been successful, it has a fundamental limitation: it scores each item independently rather than evaluating the recommendation sequence jointly.

\begin{figure}[t!]
\centering
\includegraphics[width=0.92\linewidth]{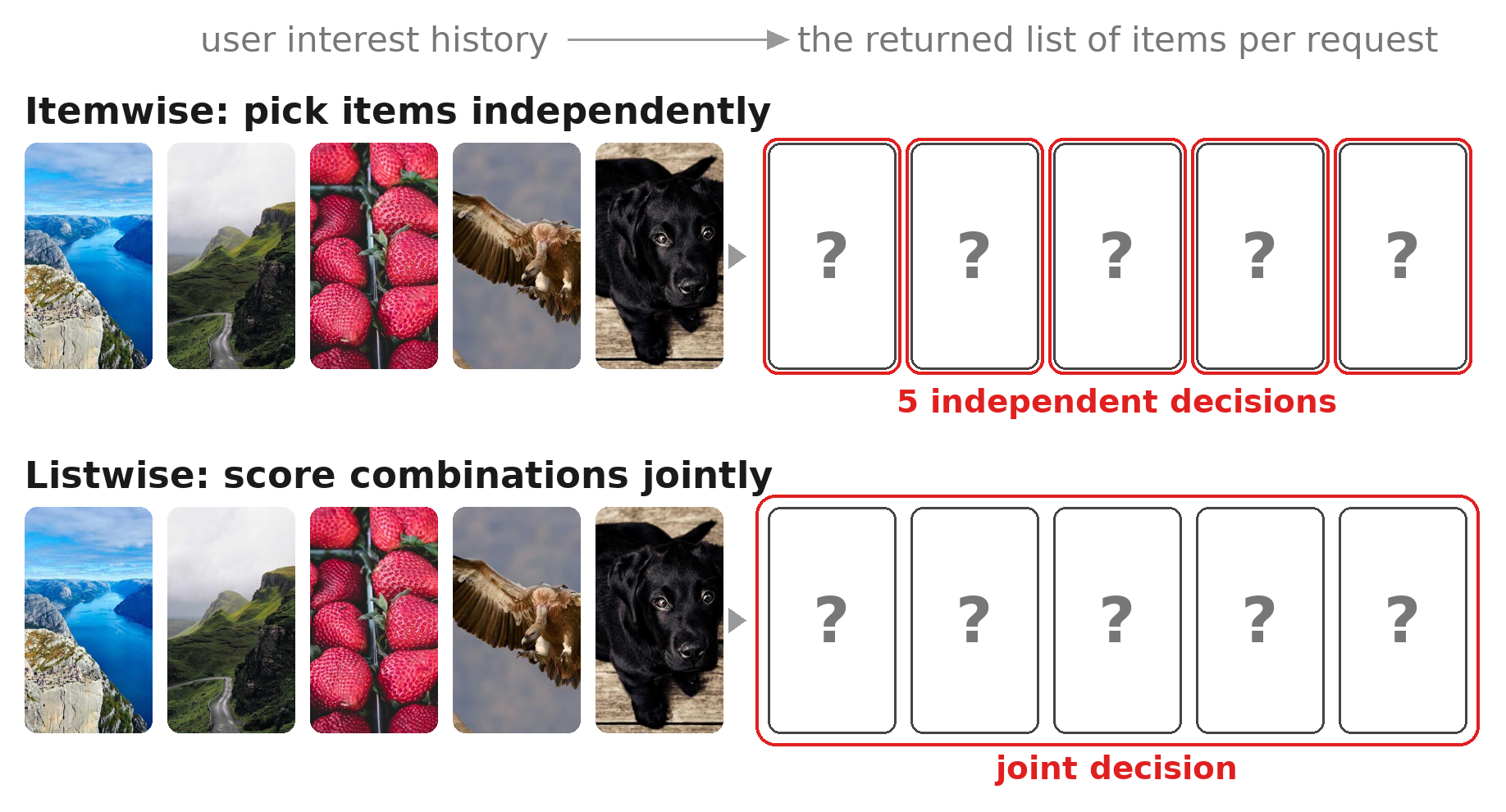}
\caption{Itemwise vs. listwise optimization. Given the user's interest history, itemwise optimization independently scores candidates and returns the top ones, whereas listwise optimization returns the best combination.}
\label{fig:idea}
\end{figure}

In contrast, LLMs fundamentally address the same problem -- generating the best sequence conditioned on a user’s request—but follow a completely different technical paradigm. Rather than independently selecting the best token at each position, an LLM generates each token conditioned on the previously generated context, with the quality of the final output determined by the sequence as a whole.
This contrast exposes the gap in paradigm we address between traditional itemwise and generative listwise recommendation:

\begin{tcolorbox}[halign=center]
{\bf Itemwise} 
(Independent Optimization)
vs. 
{\bf Listwise} 
(Joint Optimization)
\end{tcolorbox}

\begin{figure*}[t]
\centering
\includegraphics[width=\textwidth]{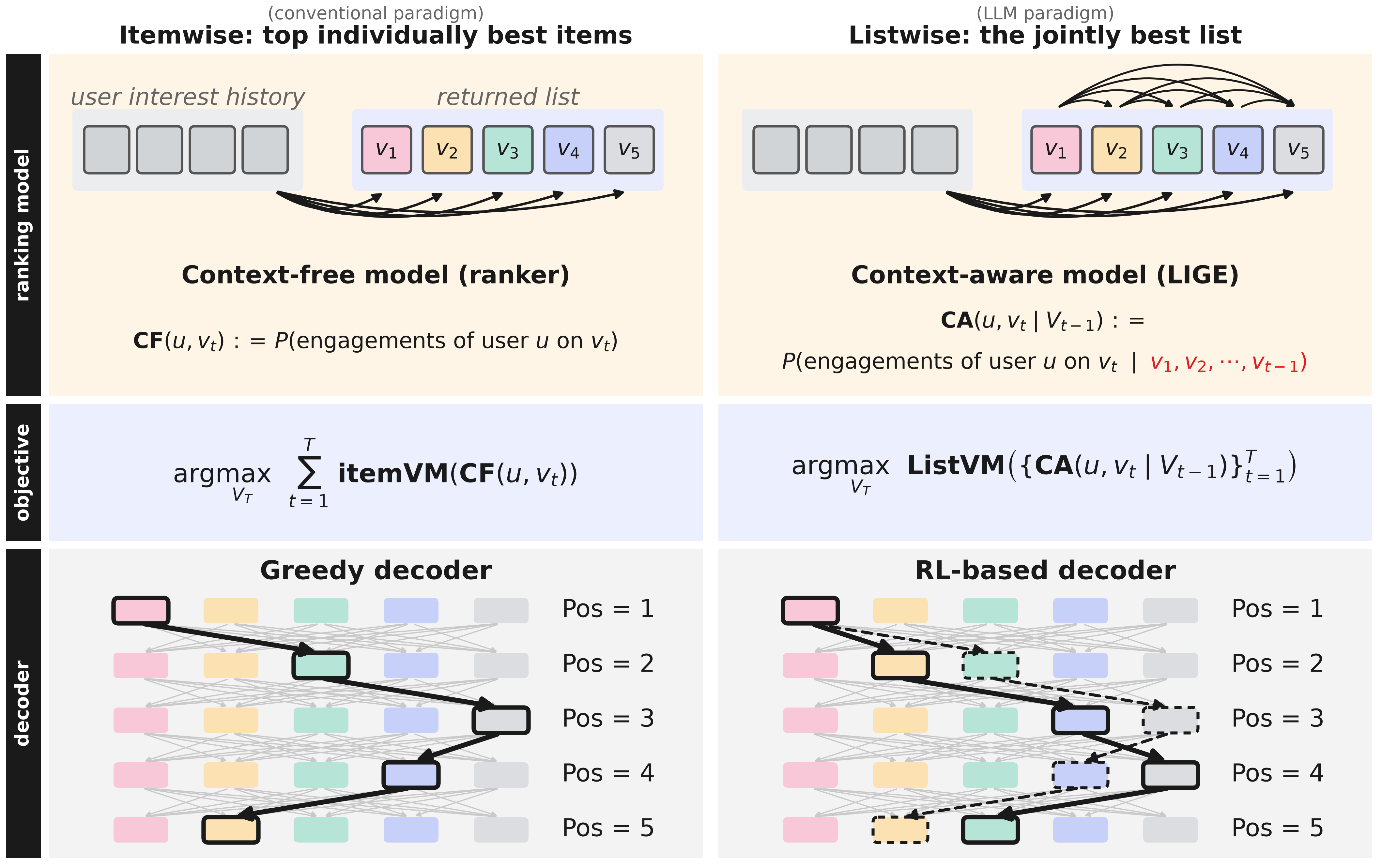}
\caption{Itemwise versus listwise recommendation across the three components $\ouralg$ upgrades. Ranking model: a context-free ranker $\textbf{CF}$ scores each item from the user alone, whereas the context-aware model $\textbf{CA}$ also conditions on the items already placed in the list (red). Objective (defined by the value model): from a sum of itemwise scores to a listwise value over the whole sequence. Decoder: incumbent itemwise greedy selection commits to a single path through the position-by-candidate lattice, whereas the RL-based decoder 
explores alternative paths 
and returns the best list found.}
\label{fig:framework}
\end{figure*}
\noindent As Figure~\ref{fig:idea} illustrates, itemwise recommendation scores items independently, whereas listwise recommendation evaluates them as a sequence. Realizing this sequential optimization requires three capabilities: predicting each candidate's value in the context of items already selected, evaluating the sequence as a whole, and searching for a feasible sequence under product and latency constraints.

Motivated by the success of LLMs, a recent trend explores achieving these capabilities by rebuilding the recommender from scratch as a fully generative system, e.g., in~\citet{onerec}. However, a successful paradigm to fully leverage listwise optimization for mature industrial applications has yet to emerge, leaving this an open problem. Specifically, attempting a wholesale replacement of an existing stack faces two additional critical barriers:
\begin{itemize}[wide=0pt, leftmargin=\parindent]
    \item {\bf System challenge. } Mature recommender systems encode years of model improvements, product logic, serving optimizations, and business constraints. A replacement can be disadvantaged in early comparisons because it must first recover much of this accumulated baseline value, therefore fair evaluation often requires careful, extended validation. Once deeply integrated, the new system can also raise rollback and reliability risks, because reverting may no longer mean disabling an isolated component.
    \item {\bf Organization challenge. } Recommendation, advertising, and search teams are typically organized around existing recommender-system components, including retrieval, ranking, value modeling, serving infrastructure, and product policy. Moving to a different paradigm can therefore disrupt not only the technical stack but also team boundaries, ownership, and long-term planning.
\end{itemize}

$\ouralg$ addresses 
both the listwise technical challenges and the replacement challenge through an {\bf additive}, {\bf revertible}, and {\bf low-resource-requirement} 
framework that generalizes the existing itemwise recommendation system. 
$\ouralg$ preserves the structure of the mature system while adding three components: 
a listwise module in the ranking model, an extension from itemwise to listwise value modeling, and an upgrade from the incumbent itemwise greedy decoder---which selects the highest-scoring remaining candidate at each list position---to an RL-based sequence decoder, as shown in Figure~\ref{fig:framework}. $\ouralg$ is thus not a replacement of traditional recommendation but a generalization 
and an upgrade.

We validated $\ouralg$ in short-video recommendation 
on \prodA{} and \prodB{}, 
both against strong, optimized baselines. 
On \prodA{}, $\ouralg$ increases 
time spent by 1.14\%, while requiring additional inference resources equivalent to roughly 10\% of those used by the context-free ranking component and increasing end-to-end per-request latency by approximately 7\% relative to the incumbent baseline. On \prodB{}, $\ouralg$ increases
time spent by 0.72\%.

\section{$\ouralg$: the Generative Paradigm}\label{sec:lige}
The section introduces the proposed generative paradigm. We start with the problem definition and the reformulation of itemwise recommendation into the listwise or generative framework, which upgrades the existing itemwise system to the listwise system. The section ends with the introduction to the listwise recommendation.

\subsection{Problem Statement}
\label{sec:problem-statement}

For each user request, the recommender system returns an ordered list of content items:
\begin{align}
{V_{T} = [v_1, v_2, \ldots, v_T], \qquad v_t \in \mathcal C.}
\label{eq:list-definition}
\end{align}
Here $\mathcal C$ is the candidate set for the request, and $V_t=[v_1,\ldots,v_t]$ denotes the selected prefix---the partial list after $t$ positions, with $V_0=\emptyset$; the items in a feasible $V_T$ are distinct. In our setting, $T$ is around 10.
The items are displayed sequentially to the user. The objective is to construct a personalized list that maximizes user experience and product quality.

\subsection{Reformulating Itemwise Recommendation into a Generative Framework}

A traditional industrial recommender system is essentially an itemwise optimization system. For each candidate item, a ranking model predicts a set of user engagement signals, such as:
\begin{align}
p_{\text{like}}, \quad p_{\text{follow}}, \quad p_{\text{share}}, \quad p_{\text{watchtime > 10s}}, \ldots
\end{align}
An itemwise value model ({\bf itemVM}) then combines these predicted engagement signals $p$ into a scalar item score. A typical item-level value model can be written as:
\begin{align}
\textbf{itemVM}(p)
=
w_1 \cdot p_{\text{like}}
+
w_2 \cdot p_{\text{follow}}
+
w_3 \cdot p_{\text{share}}
+
\cdots.
\label{eq:itemvm}
\end{align}
The system then selects the top $T$ items according to their item-level VM scores and returns them as the final recommendation list. In mature systems, the raw VM score is often further adjusted by diversity penalties, integrity rules, and other product constraints. These adjustments introduce some sequence-level awareness, but they are usually implemented as rule-based heuristics rather than learned listwise optimization.
The itemwise recommendation objective can be abstracted as:
\begin{align}
\argmax_{V_{T}} \quad
\sum_{t=1}^{T}
\textbf{itemVM}\left(\textbf{CF}(u, v_t)\right),
\label{eq:itemwise_obj_basic}
\end{align}
where \textbf{CF} denotes a \emph{context-free} predictor (the item-wise ranking model prediction of various user $u$'s engagement signals such as $p_{\text{like}}$ for item $v_t$) that scores each item independently given the user feature $u$ and the item feature $v_t$. Under this objective, the optimal list consists of the top $T$ items among the candidate set ranked by their {\bf itemVM} scores in Eq.~\eqref{eq:itemvm}.

As mentioned previously, many mature recommendation systems include an additional component---a control layer, although its name may vary---to enforce diversity in the recommended list. Its purpose is to prevent similar content, such as items from the same category, from appearing too close together. Popular approaches include gap demotion rule \citep{gong2021edge, pei2019personalized}, determinant point process (DPP) \citep{pan2020purs, meng2019tensorized, wang2021sliding, li2018fast}, and hard-coded business restrictions. {We denote the additive control-layer adjustment for candidate $v_t$ given prefix $V_{t-1}$ by $\textbf{CL}(v_t \mid V_{t-1}) \in \mathbb{R} \cup \{-\infty\}$. Finite values modify the candidate's value-model score, while $-\infty$ masks an infeasible candidate. The following examples illustrate these cases.}
\begin{itemize}[wide=0pt, leftmargin=\parindent]
\item (\textbf{hard business rule}) if the category of $v_t$ and some category in $V_{t-1}$ are not allowed to appear within the same list because of some hard restrictions, then
\[
\textbf{CL}(v_t \mid V_{t-1}) = -\infty;
\]
\item (\textbf{example gap demotion rule}) if the closest item in $V_{t-1}$ to $v_t$ belonging to the same category is at position $t' \in \{1, \cdots, t-1\}$, then
\[
\textbf{CL}(v_t \mid V_{t-1}) = - \exp(t'- t + 1)\cdot \text{constant};
\]
\item (\textbf{example DPP diversity score}) measure the incremental diversity of $v_t$ on the top of $V_{t-1}$
\[
\textbf{CL}(v_t \mid V_{t-1}) = \log \det (\Phi_t^\top \Phi_t) - \log \det(\Phi_{t-1} ^\top \Phi_{t-1}),
\]
where $\Phi_t := [\phi_t, \phi_{t-1}, \cdots, \phi_1]$ and $\phi_{t'}$ corresponds to the embedding of $v_{t'}$ normalized by $\|\phi_{t'}\| = 1$.
\end{itemize}
Adding the control-layer term to Eq.~\eqref{eq:itemwise_obj_basic} gives the overall objective:
\begin{align}
\argmax_{V_{T}} \quad
\sum_{t=1}^{T}
\left[\textbf{itemVM}\left(\textbf{CF}(u, v_t)\right) + \textbf{CL}(v_t \mid V_{t-1})\right].
\label{eq:itemwise_obj_full}
\end{align}
The incumbent itemwise system obtains the returned list with an itemwise greedy decoder 
over the candidate set $\mathcal C$:
\begin{align*}
v_{t} = \argmax_{c \in \mathcal{C}\backslash V_{t-1}}\quad \textbf{itemVM}\left(\textbf{CF}(u, c)\right) + \textbf{CL}(c \mid V_{t-1}), \; t=1, \cdots, T.
\end{align*}

\subsection{$\ouralg$: Listwise Recommendation as Generative Recommendation}

$\ouralg$ upgrades the itemwise recommendation system into a listwise generation system. In its basic vanilla form, $\ouralg$ targets the following sequence-level objective using contextual model predictions:
\begin{align}
\argmax_{V_{T}}\quad
\sum_{t=1}^{T}
\left[\textbf{itemVM}
\left(
\textbf{CA}(u, v_t \mid V_{t-1})
\right) + \textbf{CL}(v_t \mid V_{t-1})\right],
\label{eq:listVM_basic}
\end{align}
where $\textbf{CA}$ is a context-aware predictor that estimates the engagement values of item $v_t$ conditioned on the preceding items $V_{t-1}$ within the same list.

The overall upgrade from itemwise recommendation to $\ouralg$ comprises three component upgrades:
\begin{itemize}[wide=0pt, leftmargin=\parindent]
\item {\bf Ranking model:} from a context-free predictor to a context-aware predictor.
\item {\bf Value model:} from an itemwise VM to a listwise VM.
\item {\bf Decoder:} from incumbent itemwise greedy selection to RL-based sequence decoding.
\end{itemize}
Importantly, $\ouralg$ strictly generalizes the incumbent itemwise recommender. Reverting its three upgraded components to the context-free predictor, itemwise VM+CL evaluator, and itemwise greedy decoder recovers the incumbent system described above. The corresponding decoder settings are given in Section~\ref{sec:palette}.
This property matters in practice. It means $\ouralg$ can be introduced as a smooth upgrade to the existing recommender system, rather than as a disruptive replacement.

\begin{figure*}[t]
\centering
\includegraphics[width=\textwidth]{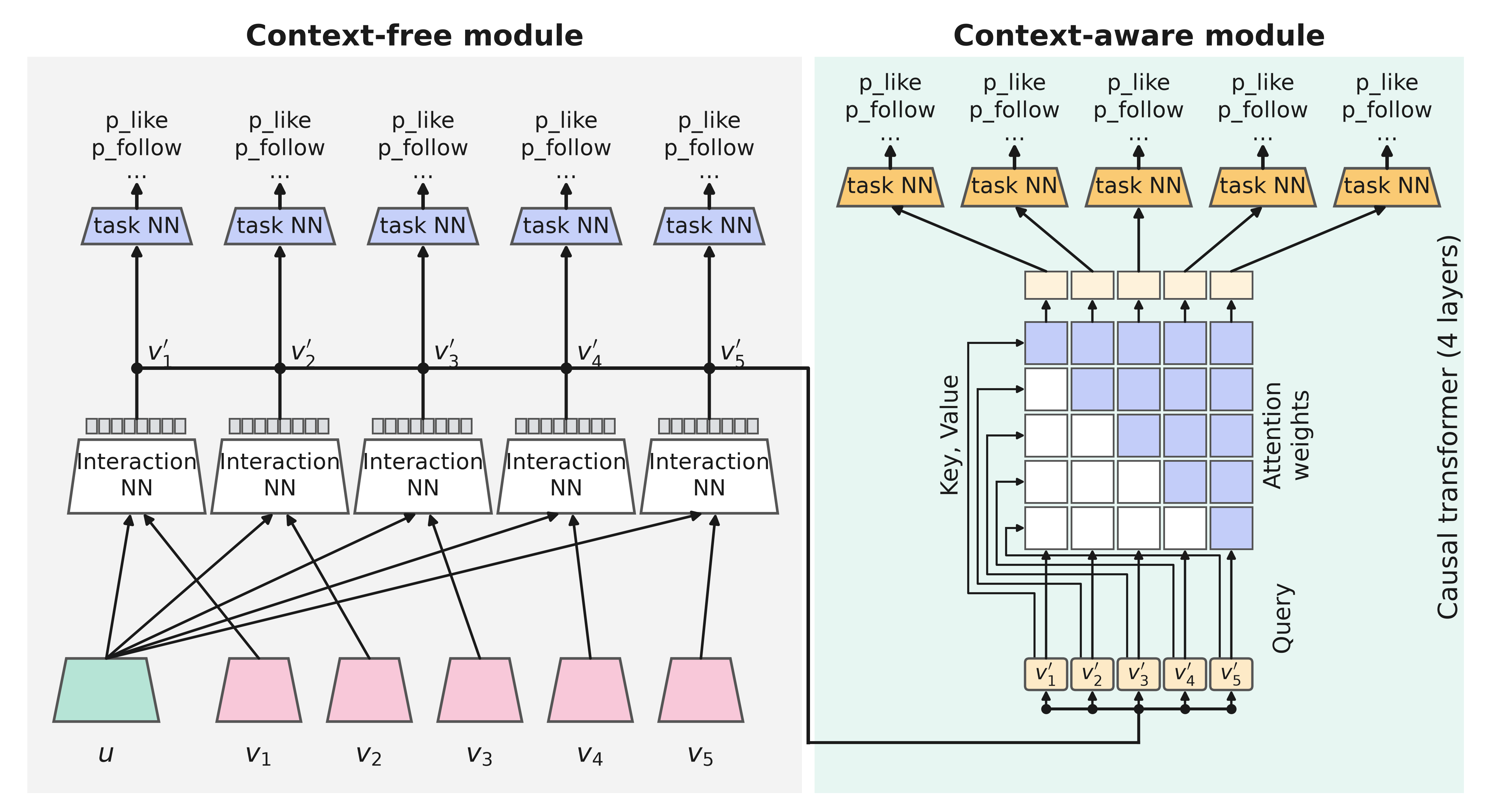}
\caption{The listwise model. The context-free module (left) is the itemwise model: each candidate $v_t$ is scored together with the user features $u$ by an interaction network and per-task heads. Its intermediate representations $v'_t$ are handed to the lightweight context-aware module (right), which refines all per-task predictions with a four-layer causal Transformer that attends only to preceding items. Within each module, all task heads share the same network weights, applied at every item or position.}
\label{fig:model-arch}
\end{figure*}

This summed objective already benefits from context-aware prediction; Section~\ref{sec:listwise-vm} upgrades it into a true listwise value model. In its most general form, $\ouralg$ targets
\begin{align}
\argmax_{V_{T}}\quad
\textbf{ListVM}
\left(\left\{
\textbf{CA}(u, v_t\mid V_{t-1})
\right\}_{t=1}^T\right).
\label{eq:listVM_generic}
\end{align}
Here $\textbf{ListVM}$ may use item metadata and includes the control-layer adjustments applied to the selected sequence.

\section{$\ouralg$ Design}
The section introduces the detailed design in $\ouralg$ in each upgraded component: the listwise model, the listwise VM, the \palette\ decoder, and serving optimization (see Appendix~\ref{app:serving}).
\subsection{Listwise Model: Context-Aware and Context-Free Predictors}

$\ouralg$ 
upgrades the existing itemwise ranking model into a listwise model. It contains two components, as illustrated in Figure~\ref{fig:model-arch}:
\begin{itemize}[wide=0pt, leftmargin=\parindent]
\item A context-free predictor, which corresponds to the existing item-wise ranking model's predictions.
\item A context-aware predictor, which refines predictions based on the previously selected items.
\end{itemize}

The context-free predictor can be written as:
\[
(u, v_t) \rightarrow y_t,\quad \forall t = 1,\cdots, T,
\]
where $u$ represents user features, $v_t$ represents the candidate item at position $t$, and $y_t$ denotes predicted engagement signals, such as like probability, follow probability, 
or watch-time prediction.

The logical form of the context-aware predictor is:
\[
(u, v_t \mid v_1, v_2, \ldots, v_{t-1}) \rightarrow y_t, \quad \forall t = 1,\cdots, T.
\]
That is, the prediction for item $v_t$ is conditioned not only on the user and the item itself, but also on the previously selected sequence within the same requested list. This allows the model to capture listwise effects such as repetition, saturation, complementarity, diversity, and user fatigue.
Because the context provides additional information, a well-designed context-aware predictor should provide more accurate predictions than a context-free predictor.

In practice, for efficiency and maintainability, $\ouralg$ does not rebuild the entire ranking model from scratch. Instead, the context-aware predictor is implemented as a lightweight refinement module on top of the existing context-free model leveraging a GPT-style decoder-only causal transformer. Specifically, let $v'_t$ denote an intermediate representation produced by the context-free model for item $v_t$. The context-aware module takes the sequence of intermediate representations as input (similar to token embeddings in the LLM):
\[
(v'_t \mid v'_1, v'_2, \ldots, v'_{t-1}) \rightarrow y_t,\quad \forall t = 1,\cdots, T.
\]

This design has two practical advantages:
\begin{itemize}[wide=0pt, leftmargin=\parindent]
\item In the evaluated \prodA{} 
configuration, the additional model remains lightweight, requiring additional inference resources equivalent to roughly 10\% of those used by the $\textbf{CF}$ component.
\item The original context-free model is fully preserved, which allows the system to reuse existing training and serving infrastructure with minimal disruption.
\end{itemize}

In our implementation, the context-aware module uses a lightweight causal GPT-like decoder-only transformer with four heads and four layers. The architecture itself is not the primary focus of this paper; future work can further optimize model design and serving efficiency.

\subsection{Listwise VM: From Itemwise Value to Listwise Value}
\label{sec:listwise-vm}

Traditional recommender systems define value at the item level. The value model estimates the utility generated by showing a single item to a user. However, user experience is inherently listwise. The value of an item depends not only on the item itself, but also on its position and on the previously consumed items.

A simple listwise VM can be constructed by summing the context-aware item values across positions:
\begin{align}\label{eq:listVM_van}
\textbf{ListVM}_{\text{vanilla}}(V_{T})
= \;&
\sum_{t=1}^{T}
\left[\,\textbf{itemVM}
\left(
\textbf{CA}(u, v_t \mid V_{t-1})\right) + \textbf{CL}(v_t \mid V_{t-1})\,\right].
\end{align}
This formulation is simple and already incorporates sequential context through the context-aware predictor. However, it ignores a fundamental factor: not every item in the list is necessarily reached by the user. Items at later positions should be weighted by the probability that the user continues watching until that position.

Therefore, $\ouralg$ introduces a more principled listwise VM based on continuation probability. Let $p_{\text{continue}}(V_{t-1})$ denote the probability that the user continues after consuming the prefix $V_{t-1}$ and reaches item $v_t$. The listwise VM is defined as follows:
\begin{align}\label{eq:listVM_golden}
\textbf{ListVM}_{\text{golden}}(V_{T})
= \;&
\sum_{t=1}^{T}
p_{\text{continue}}(V_{t-1})
\cdot
\left[\textbf{itemVM}
\left(
\textbf{CA}(u, v_t \mid V_{t-1}) \right) + \textbf{CL}(v_t \mid V_{t-1})\right].
\end{align}

The continuation probability can be recursively estimated as:
\[
p_{\text{continue}}(V_{t})
=
p_{\text{continue}}(V_{t-1})
\cdot
\textbf{CA}_{\text{continue}}(u, v_t \mid V_{t-1}),
\]
where $\textbf{CA}_{\text{continue}}(u, v_t \mid V_{t-1})$ is predicted by the context-aware model and $p_{\text{continue}}(V)$ denotes the cumulative survival of prefix $V$.

This formulation~\eqref{eq:listVM_golden} provides a statistically more faithful estimate of the expected value of the whole list, as each item’s contribution is weighted by the probability that the user actually reaches it.

\subsection{\palette\ Decoder: An RL-Based Decoder}
\label{sec:palette}

Next we introduce how to optimize the defined $\textbf{ListVM}_{\text{golden}}$ objective in \eqref{eq:listVM_golden}, given the context aware model \textbf{CA}. It also covers optimizing the $\textbf{ListVM}_{\text{vanilla}}$ objective \eqref{eq:listVM_van} by setting $p_{\text{continue}} \equiv 1$. 

Mathematically, this is an optimal sequential decision making problem with absorbing states. The proposed RL-based \palette\ decoder (as stated in Algorithm~\ref{alg:palette}) generates the output list one by one by iteratively expanding the global subsequence set $\mathcal{E}$ and only keeping top-$b$ in $\mathcal{E}$.

Specifically, \palette\ uses $\textbf{ListVM}_{\text{golden}}$ while constructing the recommendation list one position at a time. It starts from $V_0=\emptyset$. At position $t{+}1$, \palette\ expands each retained sequence $V_t$ separately by filling the next position to obtain $V_t{+}c$ where $c\in\mathcal C\setminus V_t$ allowed by the control layer. To avoid exponential computational complexity, it keeps only the top-$b$ extensions with the highest search scores. Repeating this step until $t=T$ avoids evaluating every possible full list.

In this process, the key is how to define ``top'' in expansion. In \palette\, we use the following evaluation criteria. Let $V{+}c$ be the target subsequence, its $Q$-value is calculated by
\begin{align}
Q(V{+}c)=\textbf{ListVM}_{\text{golden}}(V{+}c)+\widehat F(V{+}c).
\label{eq:Q}
\end{align}
where $\textbf{ListVM}_{\text{golden}}(V)$ evaluates a subsequence $V$, with $\textbf{ListVM}_{\text{golden}}(\emptyset)=0$. 
Algorithm~\ref{alg:palette} retains the $b$ extensions with largest $Q$. The beam width $b$ determines how many sequences are retained, while $Q$ determines which sequences are retained.

This construction admits a value-based RL interpretation: subsequence $V$ is the state, next candidate $c$ is the action, and $V{+}c$ is the successor state. The accumulated $\textbf{ListVM}_{\text{golden}}(V{+}c)$ is the realized return, while $\widehat F(V{+}c)$ is the estimated value-to-go from the successor state. Their sum $Q(V{+}c)$ is used for action selection. 
On the other hand, replacing $\textbf{CA}$ with $\textbf{CF}$, using the itemwise VM+CL evaluator, and setting $b=1$, $p_{\mathrm{continue}}\equiv1$, and $\widehat F=0$ recovers the incumbent itemwise greedy decoder.

\begin{algorithm}[t]
\caption{\palette\ Decoding}
\label{alg:palette}
\begin{algorithmic}[1]
\Require user context $u$, candidates $\mathcal{C}$, beam width $b$, list length $T$, context-aware model $\textbf{CA}$ and continuation predictor $\textbf{CA}_{\mathrm{continue}}$, item value model $\textbf{itemVM}$, control layer $\textbf{CL}$, future-value estimator $\widehat F$ (e.g., Eq.~\eqref{eq:duration-vm})
\State $\mathcal{B} \gets \{\,\emptyset\,\}$;\quad $\textbf{ListVM}_{\text{golden}}(\emptyset) \gets 0$;\quad $p_{\text{continue}}(\emptyset) \gets 1$
\Statex \textbf{Notation:} $V\in\mathcal B$ is a retained prefix; $c\in\mathcal C\setminus V$ is an unselected candidate; $V{+}c$ appends $c$ to $V$ (Eq.~\eqref{eq:list-definition}).
\Statex (To recover the incumbent itemwise decoding, use $\textbf{CF}$ in place of $\textbf{CA}$, set $\textbf{CA}_{\mathrm{continue}}\equiv1$, $b=1$, and $\widehat F=0$.)
\For{$t = 1, \ldots, T$}
  \State $\mathcal{E} \gets \{\, V{+}c \;:\; V \in \mathcal{B},\; c \in \mathcal{C} \setminus V,\; \textbf{CL}(c \mid V) > -\infty \,\}$ \Comment{admissible extensions; assume $\ge 1$ per position}
  \For{each $V{+}c \in \mathcal{E}$}
    \State $\textbf{ListVM}_{\text{golden}}(V{+}c) \gets \textbf{ListVM}_{\text{golden}}(V) + p_{\text{continue}}(V) \cdot \left[\textbf{itemVM}(\textbf{CA}(u,c\mid V))+\textbf{CL}(c\mid V)\right]$
    \State $p_{\text{continue}}(V{+}c) \gets p_{\text{continue}}(V) \cdot \textbf{CA}_{\text{continue}}(u,c\mid V)$
    \State $Q(V{+}c) \gets \textbf{ListVM}_{\text{golden}}(V{+}c) + \widehat F(V{+}c)$
  \EndFor
  \State $\mathcal{B} \gets \operatorname*{arg\,top\text{-}b}_{V' \in \mathcal{E}}\; Q(V')$ \Comment{the $b$ highest-$Q$ admissible extensions (all of $\mathcal{E}$ if $|\mathcal{E}| \le b$)}
\EndFor
\State \Return $\arg\max_{V \in \mathcal{B}} \textbf{ListVM}_{\text{golden}}(V)$ \Comment{$\widehat F(V)=0$ for full-length lists}
\end{algorithmic}
\end{algorithm}

\subsubsection{Future-Value Estimation}
The future value component $\hat{F}(\cdot)$ in \eqref{eq:Q} is the key differentiation between RL based generator and the standard beam search generator which does not have such component. However, it is also different from the standard RL based approach: \palette\ uses an estimate to drive a closed form estimate and the standard RL based approach acquires a model to estimate it.

The remaining design question is how to estimate $\widehat F(V_t)$ cheaply enough for serving. We derive a lightweight $\widehat F$ from quantities already available during decoding. Richer alternatives, such as a learned value model or a deeper planner based on Monte Carlo tree search, would require training an additional model or performing repeated model evaluations; we leave them to future work.

\paragraph{Step-based estimate.}
A first approximation uses the average per-position VM+CL score observed in the selected prefix before continuation weighting. Define
\begin{align}
\bar s(V_t)
=
\frac{1}{t}\sum_{\tau=1}^{t}
\left[\textbf{itemVM}(\textbf{CA}(u,v_i\mid V_{\tau-1}))
+\textbf{CL}(v_\tau\mid V_{\tau-1})\right].
\label{eq:average-prefix-score}
\end{align}
Assume each remaining step has the same continuation probability as the most recently selected item, $\textbf{CA}_{\text{continue}}(u,v_t\mid V_{t-1})$. Then the probability of reaching the $j$-th future position is approximated by $\textbf{CA}_{\text{continue}}(u,v_t\mid V_{t-1})^j$, giving

\begin{align}
\widehat F_{\text{step}}(V_t)
=
\bar s(V_t) \cdot
\sum_{j=1}^{T-t}\textbf{CA}_{\text{continue}}(u,v_t\mid V_{t-1})^j,
\label{eq:step-vm}
\end{align}
which is zero at $t=T$ because the sum is empty.

This step-based estimate biases search toward shorter current items. $\textbf{CA}_{\text{continue}}(u,v_t\mid V_{t-1})$ is the probability of continuing after the entire item $v_t$; for the same per-second exit propensity, a longer $v_t$ has a lower continuation probability. Reusing this whole-item probability at every future step compounds the current item's duration across all unfilled positions, reducing $\widehat F_{\text{step}}(V_t)$ and making prefixes ending in long items less likely to survive beam pruning.

\paragraph{Duration-aware estimate.}
To mitigate this repeated-duration bias, rescale the most recent continuation probability to the average duration of the selected items:
\begin{align}
\textbf{CA}_{\text{continue}}(u,v_t\mid V_{t-1})^{\bar d/d_t}.
\label{eq:duration-rescaled-continuation}
\end{align}
Here $d_t$ is the duration of the most recently selected item, $D(V_t)$ is the total duration of the $t$ selected items, and $\bar d=D(V_t)/t$ is their average duration. The exponent $\bar d/d_t$ rescales the whole-item continuation probability from duration $d_t$ to duration $\bar d$. Using this rescaled probability for each future position gives
\begin{align}
\widehat F_{\text{dur}}(V_t)
=
\bar s(V_t) \cdot
\sum_{j=1}^{T-t}
\textbf{CA}_{\text{continue}}(u,v_t\mid V_{t-1})^{j\bar d/d_t},
\label{eq:duration-vm}
\end{align}
which is again zero at $t=T$, where the sum is empty.
The duration-aware treatment keeps the score in value units while replacing the most recent item's full duration with the prefix-average duration. It therefore reduces the step estimator's preference for prefixes ending in shorter items while retaining the continuation signal. We refer to the resulting decode-time evaluator, $\textbf{ListVM}_{\text{golden}}+\widehat F_{\text{dur}}$, as the duration-aware listwise VM.

\palette\ completes the $\ouralg$ pipeline: the context-aware model predicts candidate outcomes, $\textbf{ListVM}_{\text{golden}}$ defines the value of an ordered list, and \palette\ searches for the sequence returned to the user. Section~\ref{sec:exp} evaluates the resulting end-to-end system and the contribution of its decoding configurations.

\section{Serving and Efficiency Optimization for $\ouralg$}
\label{app:serving}

This section describes how to serve $\ouralg$ and the optimizations to its additional requirements for inference resources.

\subsection{Serving for Decoding}
\label{sec:decoding}

$\ouralg$ augments the existing ranking service rather than introducing a parallel serving stack. Its request path consists of two phases (Algorithm~\ref{alg:serving}):

\begin{itemize}
\item \textbf{Context-free embedding computation.} This phase is identical to the conventional itemwise ranking pipeline. The {\bf CF} module computes the intermediate representation $v'$ (encoding both user and item information, as illustrated in Figure~\ref{fig:model-arch}) for every candidate independently of the output list. Since this computation is already performed by the existing ranking system, this phase introduces no additional inference-resource requirements.
\item \textbf{Context-aware decoding.} The decoder then constructs the output list autoregressively. At each position $t$, the context-aware module re-scores the candidate set conditioned on the previously selected prefix.
\end{itemize}

The additional inference-resource requirements are modest because the expensive computation is performed only once per request. Specifically, the context-aware decoder reuses the cached representations $v'$ produced by the context-free module, which dominates the computation of the original ranking model. Moreover, the context-aware module is intentionally lightweight: it is implemented as the four-head, four-layer causal GPT-style decoder shown in Figure~\ref{fig:model-arch}, representing only a tiny fraction of the computation of the base ranking model. Consequently, each decoding step requires only a lightweight forward pass over cached representations rather than re-running the full ranking model. Overall, listwise generation performs $T$ batched lightweight decoding passes per request. Increasing the beam width $b$ enlarges the decoding batch rather than increasing the number of sequential forward passes, while the gains from wider beams quickly saturate (Appendix~\ref{app:beam-width}).

\begin{algorithm*}[t]
\caption{$\ouralg$ serving request path}
\label{alg:serving}
\begin{algorithmic}[1]
\Require user context $u$, candidates $\mathcal{C}$, beam width $b$, list length $T$, per-request latency budget $\tau$, and the inputs of Algorithm~\ref{alg:palette} ($\textbf{CA}$, $\textbf{CA}_{\mathrm{continue}}$, $\textbf{itemVM}$, $\textbf{CL}$, $\widehat F$)
\State compute and cache $\textbf{CF}(u, c)$ and its intermediate representation $v'_c$, $\forall c \in \mathcal{C}$ \Comment{Phase 1: the unchanged forward pass}
\State $\mathcal{V}' \gets \{v'_c\}_{c \in \mathcal{C}}$ \Comment{cached once per request; no candidate re-encoding during decoding}
\State $V \gets \textsc{Palette}(u, \mathcal{C}, b, T;\, \textbf{CA}, \textbf{CA}_{\mathrm{continue}}, \textbf{itemVM}, \textbf{CL}, \widehat F)$ \Comment{Phase 2 $=$ Algorithm~\ref{alg:palette}: $T$ batched module invocations, each evaluating up to $b$ beam prefixes over $\mathcal{V}'$}
\If{Phase 2 exceeds $\tau$ ms or any $\textbf{CA}$ call fails}
  \State \Return $\textsc{Palette}(u, \mathcal{C}, 1, T;\, \textbf{CF}, 1, \textbf{itemVM}, \textbf{CL}, 0)$ \Comment{per-request fallback: the itemwise decoder; reuses the cached $\textbf{CF}$ scores, no $\textbf{CA}$ calls}
\EndIf
\State \Return $V$
\end{algorithmic}
\end{algorithm*}

\subsection{Reliability and Reversibility}
Reliability and continued iteration matter when upgrading an existing system. The $\ouralg$ framework supports configuration-level reversion without retraining:
\begin{itemize}
\item Switch $\textbf{CA}$ predictions to $\textbf{CF}$; no retraining is required.
\item Recover the incumbent itemwise decoder by setting $b=1$, $p_{\text{continue}}\equiv1$, and $\widehat F=0$ in Algorithm~\ref{alg:palette}.
\end{itemize}

The preserved context-free path makes $\ouralg$ reversible at every granularity.

Per request, if the decoding phase cannot complete within the request's latency budget $\tau$ (Algorithm~\ref{alg:serving}), the system automatically falls back to the itemwise behavior for that request. Globally, reverting to the baseline configuration is a switch rather than a migration: disabling the context-aware path---and with it the listwise VM and beam search, which are built on its predictions---recovers the itemwise system exactly, by the strict-generalization property of Section~\ref{sec:lige}. The same additivity also decouples iteration: the context-aware module can be updated independently of the base model, without touching the base model's training or publishing flow.

\subsection{Efficiency Optimization}
\label{sec:efficiency}
Section~\ref{sec:decoding} explained why the architecture is resource-efficient by design: the forward pass is reused, and each decode step is a small forward pass over cached representations. This section describes the serving-side optimizations that control the remaining resource requirements---listwise construction over a full candidate set at serving traffic would still add unnecessary computation---and reports the measured requirements of the configuration.

\textit{Restricting the re-scoring pool.} The context-aware path does not need to re-score the full candidate set. The context-free scores from the first serving phase are already a high-quality itemwise ranking, and candidates ranked far beyond the list length have low selection probability, so only the top-ranked candidates---roughly a third of the set---are passed to the context-aware module for listwise construction. This bounds the overhead of the second phase regardless of the decoding configuration: in serving benchmarks across two GPU generations, the trimmed pool raises the context-aware path's throughput by roughly 60--80\% relative to re-scoring the full set, and the online improvements of Section~\ref{sec:exp} are obtained under the trimmed pool. The context-free ranking thus acts as a learned pre-filter for the listwise stage---another way the preserved itemwise system contributes to the listwise stage.

\textit{Batching the beam.} At each decode step, all beam continuations are evaluated in one batched forward pass of the lightweight context-aware module. At the same candidate-pool size and traffic, $b=6$ requires roughly $2.1\times$ the inference resources of $b=1$, giving a derived estimate of approximately 20\% of the $\textbf{CF}$ component's resources.

\textit{Right-sizing generation.} The remaining knobs match compute to where quality still improves. Beam width is set at the saturation knee of the offline gain curve (Figure~\ref{fig:beam-width}): beyond small widths, additional beams provide little score improvement while still requiring more inference resources. In one evaluation setting, the decoder also generates only as many positions as the response actually requests---response sizes vary at serving time---rather than always decoding the maximum list length and truncating.

Across both evaluated settings, the base configuration requires additional inference resources equivalent to roughly 10\% of those used by the \textbf{CF} component. On \prodA{}, the evaluated upgrade increases end-to-end per-request latency by roughly 7\% relative to its baseline. On \prodB{}, it increases average serving latency by about 2.2\% relative to its baseline. The latency definitions and baselines differ across the two settings, so the magnitudes are not directly comparable. These measured resource and latency changes accompany the online improvements reported in Section~\ref{sec:exp}.

\input{./experiments}

\section{Related Work}
  \label{sec:related_work}
We review the related work in this section from four aspects: itemwise recommendation, traditional listwise recommendation, generative and autoregressive slate optimization, and LLM-inspired generative recommendation. The difference from prior work is that this paper addresses recommendation from an upgrade-path perspective. Instead of focusing on improving a single model or algorithm, it studies how to upgrade a mature itemwise recommendation system---refined through years of iteration---into an LLM-inspired generative recommendation paradigm. The emphasis is not only on enabling richer listwise optimization, but also on providing a practical migration path that minimizes disruption to existing infrastructure, product logic, and engineering investment.

          \noindent {\bf Itemwise Recommendation.}
    Early work on recommendation primarily follows a simple yet effective
  itemwise paradigm~\citep{wang2021dcn,wang2017deep,cheng2016wide,zhou2018deep,ma2018modeling,rendle2010factorization,ning2011sparse,sarwar2001item,guo2017deepfm}.
  Item and user features are first compressed by feature
  encoders~\citep{wang2021dcn}, and items are then
  ranked independently by their relevance to the user.
  Research along this line explores better feature modeling, user--item
  relevance scoring, and multi-task
  learning.
                      A parallel line applies sequence modeling to user interaction histories while still
  scoring each candidate independently, through self-attention~\citep{kang2018self,sun2019bert4rec},
  recurrent~\citep{hidasi2016session} and convolutional~\citep{tang2018personalized} encoders,
  target-aware attention~\citep{zhou2018deep,xia2023transact}, memory-based modeling of
  lifelong behavior~\citep{pi2019practice}, and industrial sequential
  transduction~\citep{hstu}.
      $\ouralg$ instead conditions each candidate prediction on the items already selected for the current list.
  
          \noindent {\bf Traditional Listwise Recommendation.}
     Listwise Recommendation was introduced to address the above problem~\citep{cao2007learning,ai2019learning}, while a large body of
  work introduces cross-item modeling at a dedicated \emph{reranking}
  stage.
  DLCM~\citep{ai2018learning} encodes ranking context with a GRU to
  refine scores; PRM~\citep{pei2019personalized}
  and SetRank~\citep{pang2020setrank} apply bidirectional self-attention
  over the candidate set to capture
  mutual influence in a single pass; PEAR~\citep{li2022pear} adds
  personalized contextualized transformers;
  MIR~\citep{xi2022mir} jointly models set-level candidates and user
  history; and PIER~\citep{shen2023pier}
  selects among candidate permutations end-to-end.
    A prominent instantiation is the ``generator--evaluator'' (G--E)
  framework, which pairs a list generator
  with a list evaluator that scores the quality of a generated sequence~\citep{wang2019sequential,feng2021grn,feng2021revisit,ren2024nonautoregressive}. The evaluator
  guides the generator so that together they produce a high-quality list
  in terms of relevance or engagement.
  NAR4Rec~\citep{ren2024nonautoregressive} in particular constructs
  multiple slates and selects the best via a
  learned slatewise evaluator.
    Score-refinement methods form the final slate by sorting contextualized scores,
  whereas generator--evaluator methods generate candidate slates and select among
  them. Both families are typically introduced through dedicated reranking
  machinery on top of the existing system. $\ouralg$ delivers the G--E benefits
  \emph{inside} the existing ranking stage, without adding a dedicated
  reranking stage.
    Within this framing, $\ouralg$ interleaves generation and evaluation: it scores
  partial lists and prunes low-valued prefixes before completion. This contrasts
  with generate-then-evaluate variants that evaluate completed candidate slates.

          \noindent {\bf Generative and Autoregressive Slate Construction.}
    Rather than re-sorting scores, another family directly generates the
  slate. List-CVAE~\citep{jiang2018beyond}
  uses a conditional variational auto-encoder to generate stochastic
  lists, and pivot-CVAE~\citep{liu2021variation}
  improves it to guarantee list variation and mitigate
  over-concentration. GFN4Rec~\citep{liu2023generative}
  adapts GFlowNet~\citep{bengio2021flow} to sample sequences with
  probability proportional to the listwise
  reward.
    A closely related thread constructs the slate \emph{autoregressively}.
  Seq2Slate~\citep{bello2018seq2slate} selects items one at a time with
  an
  RNN encoder--decoder, compressing prior context into a fixed-size
  hidden state. SlateQ~\citep{ie2019slateq}
  decomposes the combinatorial slate reward into per-item Q-values under
  a user choice model, but does not
  model inter-item interactions during construction.
  GFN4Rec~\citep{liu2023generative} also decodes
  autoregressively but without direct attention over previously selected
  items. More recent work accelerates
  or reshapes this decoding. GReF~\citep{lin2025gref} uses ordered
  multi-token prediction and replaces the
  evaluator with preference-based training, and HiGR~\citep{pang2025higr}
  adds hierarchical planning with
  multi-objective preference alignment. Alternative generative paradigms
  avoid sequential decoding entirely.~\citet{tomasi2025diffslate} cast slate construction as
  parallel denoising diffusion, which
  removes sequential latency but cannot enforce per-step constraints.
    $\ouralg$ also adopts the autoregressive formulation, but unlike
  Seq2Slate's fixed hidden state or GFN4Rec's
  attention-free decoding, it applies full causal attention over all
  previously selected items within a
  single architecture.  A practical advantage is that $\ouralg$ has  an explicit objective, and  generation can be controlled by changing objective weights or VM forms.

          \noindent {\bf LLM-Inspired Generative Recommendation}
    Following the success of LLMs, the recommendation community has
  increasingly adopted their
  techniques in ranking systems~\citep{wu2024survey}. One idea
  is to \emph{tokenize}
  items so that recommendation can be cast as sequence generation. At
  the retrieval stage,
  TIGER~\citep{tiger}, building on the differentiable search
  index~\citep{tay2022dsi}, encodes
  each item into multi-modal semantic IDs and autoregressively predicts
  the next ID to be
  consumed, which is effectively a generative retrieval paradigm. In a similar
  spirit, streaming
  VQ~\citep{streamingvq} tokenizes items with VQ-VAE~\citep{vqvae},
  upgrading the industrial
  index into a learnable, instantly updatable, and balanced structure,
  while
  P5~\citep{geng2022p5} unifies diverse recommendation tasks under a
  single text-to-text
  formulation. Pushing this direction further, OneRec~\citep{onerec}
  couples generative
  retrieval with an Iterative Preference Alignment module and is the
  first approach to replace
  the entire recommendation funnel with one unified model.     Unlike OneRec's wholesale replacement of the funnel, $\ouralg$ reaches
  listwise, LLM-style
  generation by upgrading the existing ranking stage \emph{in place}: it
  captures contextual
  cues, makes listwise predictions, and delivers the
  generator--evaluator benefits without
  adding a stage or replacing the stack.

\section{Conclusion and Future Work}
In this paper, we present a practical and low-resource-requirement framework for upgrading traditional itemwise recommendation systems toward a generative, listwise recommendation paradigm. The proposed approach introduces a context-aware module on top of existing ranking models, upgrades the itemwise value model into a listwise value model, and replaces conventional itemwise greedy decoding with an RL-based decoder. Through validation on Instagram Reels and Facebook Video, we demonstrate the framework in real-world industrial recommendation systems.
A design goal of this work is to enable a smooth and incremental transition from mature itemwise systems to listwise generative systems. Rather than requiring a disruptive replacement of existing infrastructure, models, or organizational ownership, the proposed framework generalizes and extends the existing recommendation stack. This makes the approach more practical for industrial environments, where technical migration effort, system reliability, latency constraints, and cross-team ownership all matter.

This work should not be viewed as the final form of generative recommendation. Instead, it provides a transition framework  with headroom. For example, more expressive architectures may further improve the model’s ability to capture list-level dependencies; the listwise objective can be extended to incorporate richer list-level signals that are difficult to define in an itemwise system;
more advanced decoding and reinforcement learning methods may further improve the diversity and long-term quality of generated recommendation lists. The current $\ouralg$ framework can only handle an input candidate pool on the order of hundreds. To achieve truly end-to-end recommendation, it will need to be integrated with technologies such as Semantic IDs that can efficiently support much larger candidate spaces.

$\ouralg$ offers additional future potential for end-to-end optimization across the full serving stack. In latency-sensitive serving applications, with streaming inference, we can stream out individual results and serve them to the user as they become available. We can further optimize the case of greedy search by immediately yielding the first result before applying the CA pass, as the first result will not be affected. The complexity of the CA pass also gives additional potential for dynamic compute complexity scaling based on varying compute supply and demand.
 This work provides a foundation for future research and system development toward more expressive, controllable, and value-aligned recommendation systems.

\section*{Acknowledgments}
We thank Rex Cheung, Erica Li, Lars Backstrom, Max Eulenstein, Bruce Deng, Yimin Tan, Qichao Que, Jerry Fu, Congle Zhang, Lihong Li, Fei Sha, Zhenyu Su, and Mahesh Srinivasan for their insightful technical discussions, constructive feedback, and many valuable contributions throughout this work. We are particularly grateful to Shilin Ding for championing this project from its inception and providing steadfast support throughout its 0-to-1 journey.

\bibliographystyle{assets/plainnat}
\bibliography{ref}

\clearpage
\newpage
\beginappendix

\input{./appendix}

\end{document}

%% file: experiments.tex
\section{Experimental Results}
\label{sec:exp}
This section evaluates $\ouralg$ in short-video recommendation settings on \prodA{} and \prodB{}. Section~\ref{sec:exp_model} compares context-aware prediction ($\textbf{CA}$) with the context-free baseline ($\textbf{CF}$) using normalized entropy (NE)~\citep{liu2023learning}. Sections~\ref{sec:exp_online} and~\ref{sec:exp_upgrade} report online results for the $b=1$ base configuration against the corresponding incumbent systems and for the $b=6$ duration-aware configuration against the $b=1$ base configuration, respectively. Section~\ref{sec:ecosystem} provides additional content-ecosystem analysis. In the evaluated settings, the candidate-pool size $|\mathcal C|$ is on the order of $10^2$, while the output-list length $T$ is on the order of $10$ (Eq.~\eqref{eq:list-definition}).

\subsection{Context-Aware vs. Context-Free Prediction}
\label{sec:exp_model}

Before evaluating end-to-end ranking outcomes, we isolate the effect of contextualization on prediction quality. Table~\ref{tab:offline-model-validation} compares $\textbf{CA}$ with $\textbf{CF}$; positive values indicate that $\textbf{CA}$ achieves lower (better) NE. On \prodA{}, $\textbf{CA}$ improves all six displayed task families, led by Continue (1.57\%). On \prodB{}, it improves all five task families shown. Beyond this displayed subset, the complete 17-task \prodB{} evaluation shows prediction-quality improvements on 15 tasks and regressions of at most 0.07\% on two minor tasks. These results establish the prediction-quality benefit of $\textbf{CA}$; Sections~\ref{sec:exp_online} and~\ref{sec:exp_upgrade} next evaluate whether that benefit translates into online ranking outcomes.

\begin{table}[t]
\centering
\caption{Relative NE improvement of the context-aware predictor ($\textbf{CA}$) against the context-free baseline ($\textbf{CF}$). Positive values indicate lower (better) NE. \prodA{} values are accumulated over one online-training run; \prodB{} values come from its context-aware-versus-context-free evaluation. `---' marks a task not tracked.}
\begin{tabular}{l c c}
\toprule
Task family & \prodA{} & \prodB{} \\
\midrule
Continue & 1.57\% & --- \\
Skip & 0.74\% & 0.46\% \\
Like & 0.29\% & 0.46\% \\
Comment & 0.21\% & 1.59\% \\
Share & 0.30\% & 1.25\% \\
Watch completion & 0.43\% & 0.55\% \\
\bottomrule
\end{tabular}
\label{tab:offline-model-validation}
\end{table}

\subsection{Online Validation: $b=1$ Base Configuration}
\label{sec:exp_online}

\begin{table}[t]
\centering
\caption{
Online A/B results for the $b=1$ base configuration in Eq.~\eqref{eq:listVM_van}, each against the corresponding incumbent recommendation baseline. Sessions and DAU reflect overall product activity; the remaining metrics are restricted to the evaluated short-video surface. The \prodA{} entry in the Likes / reactions row is likes; the \prodB{} counterpart is reactions. $\dagger$ denotes $p<0.001$.
}
\begin{tabularx}{\columnwidth}{>{\raggedright\arraybackslash}X *{2}{>{\centering\arraybackslash}X}}
\toprule
Metric & \prodA{} & \prodB{} \\
\midrule
Sessions & +0.11\%$^\dagger$ & +0.07\% \\
DAU & +0.05\%$^\dagger$ & +0.01\% \\
Time spent & +1.14\%$^\dagger$ & +0.72\%$^\dagger$ \\
Video views (VPVs) & +2.28\%$^\dagger$ & $-0.52\%^{\dagger}$ \\
Likes / reactions & +2.65\%$^\dagger$ & +1.59\%$^\dagger$ \\
Reshares & +1.77\%$^\dagger$ & +0.99\% \\
\bottomrule
\end{tabularx}
\label{tab:online-ab}
\end{table}

\begin{table}[t]
\centering
\caption{Online \prodA{} A/B contrasts against the $b=1$ base configuration used in Table~\ref{tab:online-ab}. All configurations use $\textbf{CA}$, and each treatment row is compared independently with that base configuration; `---' marks the reference row. $\dagger$ denotes $p<0.001$.}
\begin{tabular}{l l c c c c}
\toprule
Objective & Decoder & \shortstack{Time spent} & \shortstack{Video views} & Likes & Reshares \\
\midrule
$\textbf{ListVM}_{\text{vanilla}}$ & $b=1,\ \widehat F=0$ & --- & --- & --- & --- \\
$\textbf{ListVM}_{\text{vanilla}}$ & $b=6,\ \widehat F=0$ & $+0.05\%$ & $+0.00\%$ & $+0.74\%^{\dagger}$ & $+1.21\%^{\dagger}$ \\
$\textbf{ListVM}_{\text{golden}}$ & $b=6,\ \widehat F=\widehat F_{\mathrm{dur}}$ & $+0.69\%^{\dagger}$ & $+1.82\%^{\dagger}$ & $+2.93\%^{\dagger}$ & $+1.41\%^{\dagger}$ \\
\bottomrule
\end{tabular}\label{tab:palette-online}
\end{table}

\begin{table*}[t]
\centering
\caption{\prodA{} paired-request list-composition changes relative to the baseline. Topic ranges span the taxonomies analyzed; \textcolor{improvegreen}{green} marks diagnostic improvements and \textcolor{regressred}{red} marks the freshness regression.}
\setlength{\tabcolsep}{4pt}
\renewcommand{\arraystretch}{0.9}
\begin{tabular}{l l c l}
\toprule
Dimension & Metric & Change & Interpretation \\
\midrule
Topic mix & Topic entropy & \impchg{+1.14\% to +2.37\%} & more topic variety \\
Topic mix & Distinct topic clusters & \impchg{+1.34\% to +2.40\%} & broader topic coverage \\
Repetition & Longest same-topic streak & \impchg{-4.03\% to -4.74\%} & shorter repeated-topic runs \\
Local similarity & Adjacent-video cosine similarity & \impchg{-3.68\%} & less adjacent similarity \\
Creator mix & Distinct creators & \impchg{+0.66\%} & more creators per request \\
Creator mix & Same-creator concentration & \impchg{-2.98\%} & less creator concentration \\
Creator mix & Creator entropy & \impchg{+0.54\%} & more balanced creator mix \\
Creator size & Large-creator prevalence & \impchg{-1.15\%} & shift away from largest creators \\
Creator size & Medium-creator prevalence & \impchg{+0.38\%} & shift toward medium creators \\
Exploration & Familiar-creator content & \impchg{-6.23\%} & more exposure beyond familiar creators \\
Exploration & Interest-matched-topic content & \impchg{-1.65\%} & more exploration beyond known interests \\
Length mix & Within-request length entropy & \impchg{+0.86\%} & more length variety \\
Freshness & Videos under 72 hours old & \regchg{-0.66\%} & fewer very-fresh videos \\
\bottomrule
\end{tabular}
\label{tab:case-study-metrics}
\end{table*}
We first test whether the prediction-quality gains above translate into online gains when $\ouralg$ is integrated into the ranking system. The base configuration is
\begin{center}
\begin{tabular}{|c|c|c|}
\hline
\textbf{Model} & \textbf{Objective} & \textbf{Decoder} \\
\hline
$\textbf{CA}$ & $\textbf{ListVM}_{\text{vanilla}}$ & $b=1,\ \widehat F=0$ \\
\hline
\end{tabular}
\end{center}
where $\textbf{ListVM}_{\text{vanilla}}$ is defined in Eq.~\eqref{eq:listVM_van}. Each product is evaluated against its own actively optimized incumbent baseline.

For \prodA{}, the treatment and baseline groups each included 1.5\% of users. For \prodB{}, the treatment group included approximately 2\% of users and was compared with a similarly sized control group. Table~\ref{tab:online-ab} reports seven-day readouts for both products. Across both evaluations, $\ouralg$ broadly improves key consumption and engagement metrics, with the individual movements and tradeoffs reported in the table. Across both settings, the base configuration requires additional inference resources equivalent to roughly 10\% of those used by the $\textbf{CF}$ component. In the \prodA{} evaluation, end-to-end per-request latency increases by roughly 7\% relative to its baseline. In the \prodB{} evaluation, average serving latency increases by about 2.2\% relative to its baseline. Because the latency definitions and baselines differ, these measurements are not directly comparable. Appendix~\ref{app:serving} provides serving and efficiency details.

These results establish online gains for the $b=1$ base configuration. We next ask whether richer decoding adds value beyond this configuration, first by widening the beam and then by enabling the combined listwise, duration-aware configuration.

\subsection{\prodA{} Online Validation: $b=6$ Duration-Aware Configuration}
\label{sec:exp_upgrade}
We evaluate this next step on \prodA{}, using the $b=1$ base configuration from Section~\ref{sec:exp_online} as the common baseline. The duration-aware configuration uses
\begin{center}
\begin{tabular}{|c|c|c|}
\hline
\textbf{Model} & \textbf{Objective} & \textbf{Decoder} \\
\hline
$\textbf{CA}$ & $\textbf{ListVM}_{\text{golden}}$ & $b=6,\ \widehat F=\widehat F_{\mathrm{dur}}$ \\
\hline
\end{tabular}
\end{center}
Here $\textbf{ListVM}_{\text{golden}}$ and $\widehat F_{\mathrm{dur}}$ are defined in Eqs.~\eqref{eq:listVM_golden} and~\eqref{eq:duration-vm}, respectively. We compare three configurations with identical weights inside $\textbf{itemVM}$. The baseline uses $b=1$, $\textbf{ListVM}_{\text{vanilla}}$, and $\widehat F=0$. The beam-width arm increases $b$ from 1 to 6 while keeping the objective and future-value estimate unchanged. The duration-aware arm uses $b=6$ with $\textbf{ListVM}_{\text{golden}}$ and $\widehat F_{\mathrm{dur}}$. Appendix~\ref{app:beam-width} motivates the choice of $b=6$.

Each $b=6$ treatment group included approximately 3\% of users and was compared with a similarly sized $b=1$ control group. Table~\ref{tab:palette-online} reports independent seven-day A/B contrasts of each $b=6$ treatment against the \prodA{} $b=1$ base configuration used in Table~\ref{tab:online-ab}. The beam-width arm primarily improves likes and reshares, while the duration-aware arm improves all four reported metrics. At the same candidate-pool size and traffic, the $b=6$ configuration is estimated to require additional inference resources equivalent to approximately 20\% of those used by the $\textbf{CF}$ component. A comparable end-to-end latency estimate is not available for $b=6$; Appendix~\ref{app:serving} provides details.

\subsection{Impact on the Ecosystem}\label{sec:ecosystem}
Beyond online outcomes, we examine how $\ouralg$ in the setting of Section~\ref{sec:exp_online} changes the ecosystem on \prodA{}. Table~\ref{tab:case-study-metrics} summarizes paired-request diagnostics across topics, creators, exploration, length, and freshness. The overall observation is $\ouralg$ improves the product ecosystem from multiple dimensions with minor regression on content freshness. \begin{figure*}[t]
  \centering
  \includegraphics[width=\textwidth]{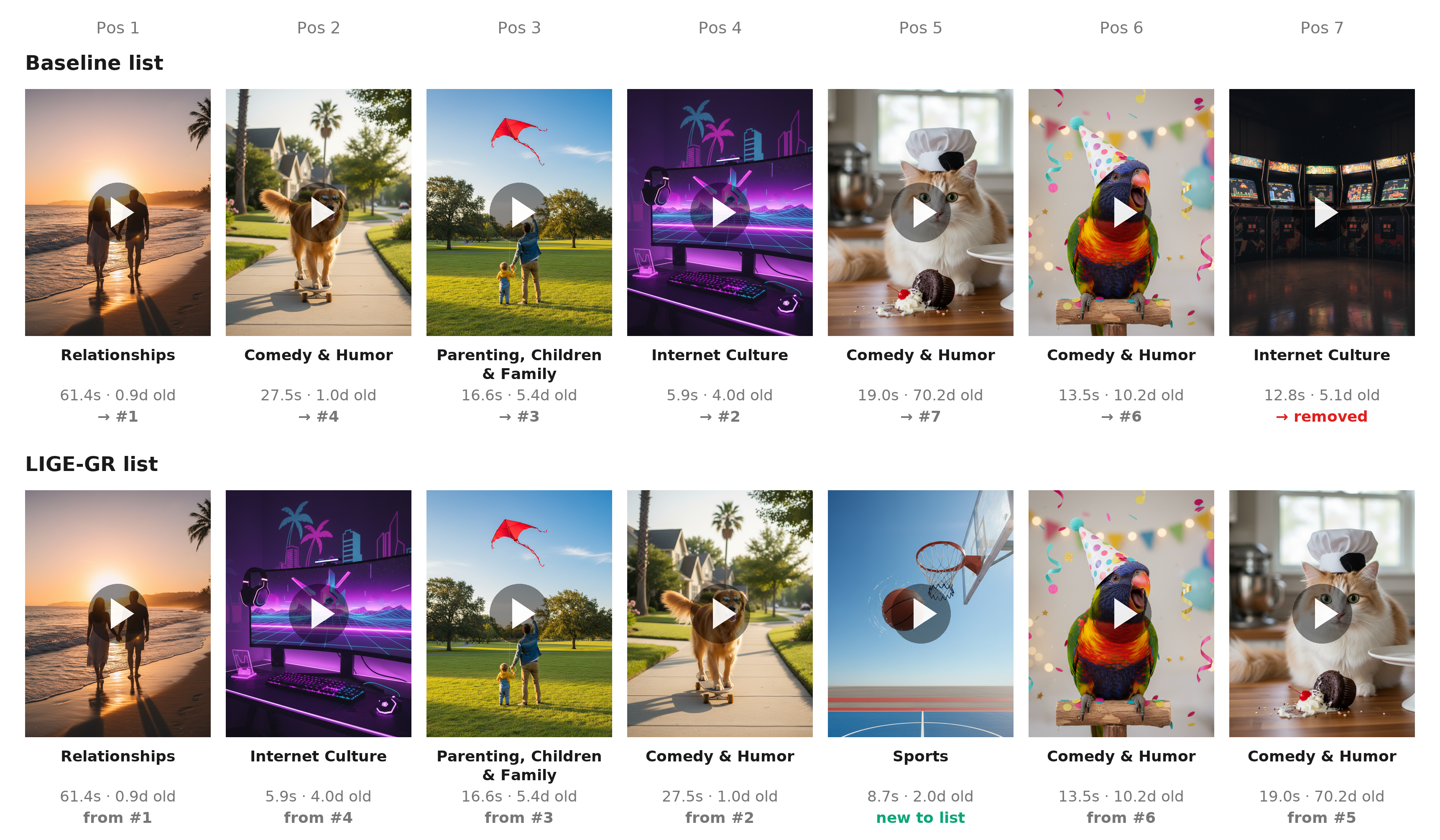}
              \caption{A representative paired request from the \prodA{} evaluation's counterfactual logs: the baseline list (top) and the $\ouralg$ list (bottom) generated from identical inputs, with each item's destination or origin position annotated. Video thumbnails are AI-generated stand-in images matching each item's topic (one per unique video); topic, duration, age, and movement annotations are from the logged request. $\ouralg$ preserves every baseline topic while introducing a new one (Sports), moves the 27.5-second video from position 2 to position 4, demotes the 70-day-old video to the final position, and removes one of the two Internet Culture items.}
  \label{fig:case-study-list-composition}
\end{figure*}

\paragraph{Methodology.}
To understand how $\ouralg$ changes generated lists, we use counterfactual logging that records both the baseline pointwise list and the $\ouralg$ context-aware list for the same request over identical inputs. We analyze 13,197 paired requests from 7,110 known logged users. 1,498 of these requests lacked logged viewer IDs and are conservatively counted as one user each, giving 8,608 effective users. These data cover a three-day window and are sampled at the request level within an experiment cohort, which skews the sample toward more-active viewers.
Table~\ref{tab:case-study-metrics} in the main text summarizes the resulting composition shifts. All reported effects are assessed to be statistically significant at the 95\% level via separate Bayesian hierarchical linear models for each estimate that account for multiple requests by the same user. We report effect sizes as relative changes compared to the baseline and interpret the main tradeoffs below.

\paragraph{Diversity and repetition.}
Table~\ref{tab:case-study-metrics} shows that $\ouralg$ increases topic variety and topic coverage while reducing repeated-topic runs. Adjacent videos also become less similar to one another on average, improving list-level variety while introducing a smoothness--diversity tradeoff that is invisible in purely pointwise evaluation.

\paragraph{Creator mix and exploration.}
The generated lists include more distinct creators, lower same-creator concentration, and higher creator entropy. They also shift away from the largest creators and from content already familiar to the user or tightly matched to known interests. We interpret these movements as an exploration tradeoff rather than an unqualified win: $\ouralg$ exposes a broader set of creators and topics, while giving up some immediate affinity matching.

\paragraph{Length and freshness.}
Within-request length entropy increases, so users see a wider variety of video lengths within the same request. The fraction of very fresh videos decreases, matching the online recency guardrail movement; we therefore treat freshness as the table's composition regression rather than as a quality improvement.

Together, Figure~\ref{fig:case-study-list-composition} and Table~\ref{tab:case-study-metrics} show that $\ouralg$ changes ecosystem exposure across topics, creators, familiarity, length, and freshness. These diagnostics characterize how generated lists redistribute exposure across the content and creators represented in those lists, while the online experiments above measure the corresponding product impact.

%% file: appendix.tex
\section{Beam-Width Replay Analysis}
\label{app:beam-width}

\begin{figure}[t]
\centering
\includegraphics[width=0.8\linewidth]{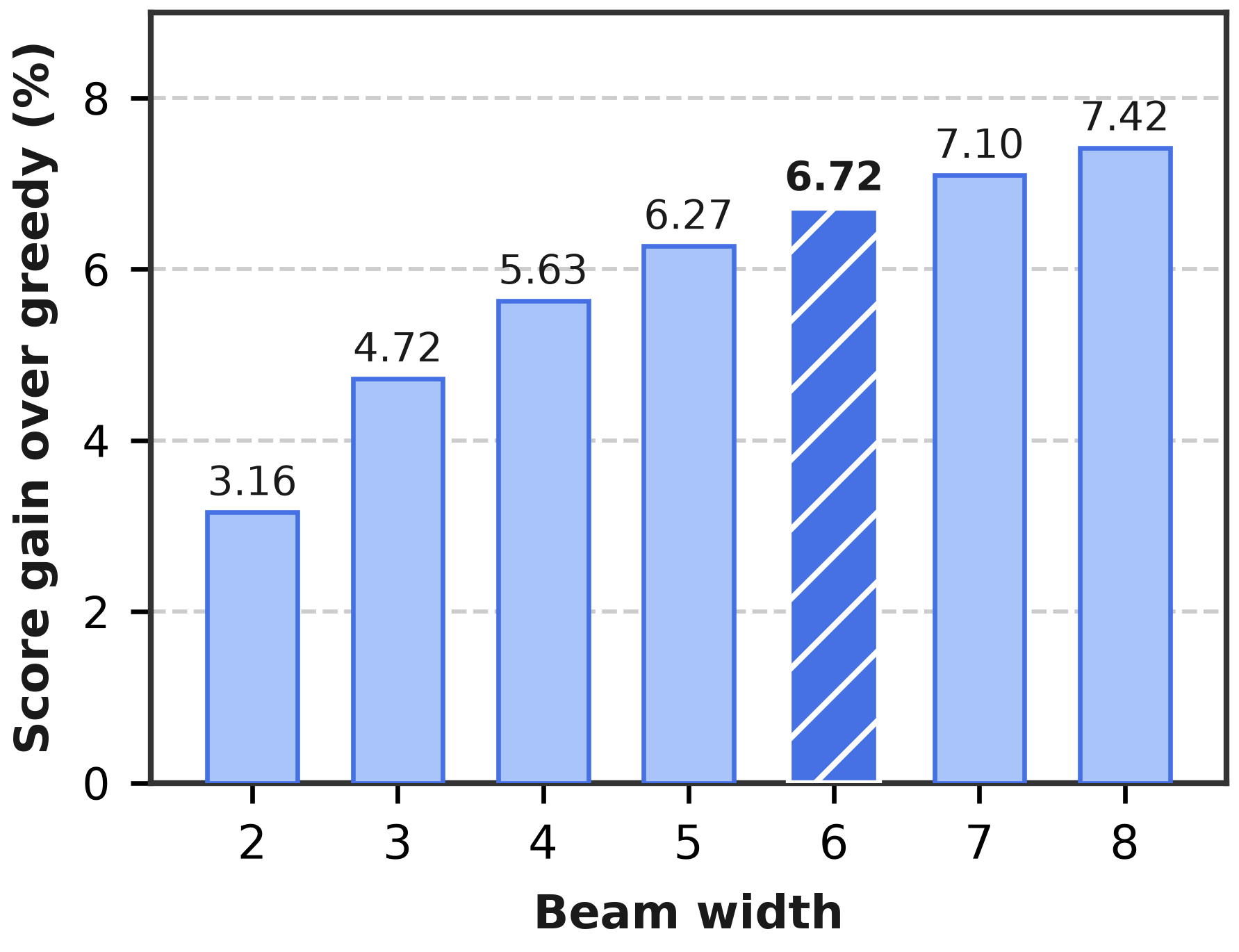}
\caption{Beam-width selection on 5{,}815 outlier-filtered replayed requests. Relative accumulated VM+CL score improvement over the $b=1$ reference, with $p_{\mathrm{continue}}\equiv1$, $\widehat F=0$, and only $b$ varying. Beam width $b=6$ lies in the saturated regime and is used in the final two configurations of Table~\ref{tab:palette-online}.}
\label{fig:beam-width}
\end{figure}

\paragraph{Set $b=6$ under accumulated VM+CL scoring.}
To choose $b$, we vary it while holding the accumulated VM+CL score fixed, with $p_{\mathrm{continue}}\equiv1$ and $\widehat F=0$. The $b=1$ equivalence check uses 11{,}965 replayed requests and reproduces the incumbent itemwise greedy decoder's mean and quartile score statistics to within 0.15\%. In a separate outlier-filtered replay of 5{,}815 requests, Figure~\ref{fig:beam-width} shows diminishing returns beyond small beam widths; the two $b=6$ configurations in Table~\ref{tab:palette-online} use $b=6$.

Increasing $b$ from 1 to 6 under the accumulated VM+CL score yields narrow online improvements: time spent and video views remain near-neutral, while likes rise by $+0.74\%$ and reshares by $+1.21\%$ relative to the matched $b=1$ configuration (Table~\ref{tab:palette-online}). The wider beam improves reaction metrics but does not by itself convert the offline score improvement into broad consumption improvements.

%% file: arxiv_main.bbl
\begin{thebibliography}{50}
\providecommand{\natexlab}[1]{#1}
\providecommand{\url}[1]{\texttt{#1}}
\expandafter\ifx\csname urlstyle\endcsname\relax
  \providecommand{\doi}[1]{doi: #1}\else
  \providecommand{\doi}{doi: \begingroup \urlstyle{rm}\Url}\fi

\bibitem[Ai et~al.(2018)Ai, Bi, Guo, and Croft]{ai2018learning}
Qingyao Ai, Keping Bi, Jiafeng Guo, and W~Bruce Croft.
\newblock Learning a deep listwise context model for ranking refinement.
\newblock In \emph{Proc.~SIGIR'18}, 2018.

\bibitem[Ai et~al.(2019)Ai, Wang, Golbandi, Bendersky, and
  Najork]{ai2019learning}
Qingyao Ai, Xuanhui Wang, Nadav Golbandi, Mike Bendersky, and Marc Najork.
\newblock Learning groupwise scoring functions using deep neural networks.
\newblock In \emph{Proc.~International Workshop On Deep Matching In Practical
  Applications'19}, 2019.

\bibitem[Bello et~al.(2018)Bello, Kulkarni, Jain, Boutilier, Chi, Eban, Luo,
  Mackey, and Meshi]{bello2018seq2slate}
Irwan Bello, Sayali Kulkarni, Sagar Jain, Craig Boutilier, Ed~Chi, Elad Eban,
  Xiyang Luo, Alan Mackey, and Ofer Meshi.
\newblock Seq2slate: Re-ranking and slate optimization with rnns.
\newblock \emph{arXiv preprint arXiv:1810.02019}, 2018.

\bibitem[Bengio et~al.(2021)Bengio, Jain, Korablyov, Precup, and
  Bengio]{bengio2021flow}
Emmanuel Bengio, Moksh Jain, Maksym Korablyov, Doina Precup, and Yoshua Bengio.
\newblock Flow network based generative models for non-iterative diverse
  candidate generation.
\newblock In \emph{Proc.~NeurIPS'21}, 2021.

\bibitem[Bin et~al.(2025)Bin, Cui, Yan, Zhao, Han, Yan, Zhang, Zhou, Yang, and
  Liu]{streamingvq}
Xingyan Bin, Jianfei Cui, Wujie Yan, Zhichen Zhao, Xintian Han, Chongyang Yan,
  Feng Zhang, Xun Zhou, Xiao Yang, and Zuotao Liu.
\newblock Real-time indexing for large-scale recommendation by streaming vector
  quantization retriever.
\newblock In \emph{Proc.~KDD'25}, 2025.

\bibitem[Cao et~al.(2007)Cao, Qin, Liu, Tsai, and Li]{cao2007learning}
Zhe Cao, Tao Qin, Tie-Yan Liu, Ming-Feng Tsai, and Hang Li.
\newblock Learning to rank: from pairwise approach to listwise approach.
\newblock In \emph{Proc.~ICML'07}, 2007.

\bibitem[Cheng et~al.(2016)Cheng, Koc, Harmsen, Shaked, Chandra, Aradhye,
  Anderson, Corrado, Chai, Ispir, et~al.]{cheng2016wide}
Heng-Tze Cheng, Levent Koc, Jeremiah Harmsen, Tal Shaked, Tushar Chandra,
  Hrishi Aradhye, Glen Anderson, Greg Corrado, Wei Chai, Mustafa Ispir, et~al.
\newblock Wide \& deep learning for recommender systems.
\newblock In \emph{Proc.~Workshop on Deep Learning for Recommender Systems'16},
  2016.

\bibitem[Deng et~al.(2025)Deng, Wang, Cai, Ren, Hu, Ding, Luo, and
  Zhou]{onerec}
Jiaxin Deng, Shiyao Wang, Kuo Cai, Lejian Ren, Qigen Hu, Weifeng Ding, Qiang
  Luo, and Guorui Zhou.
\newblock Onerec: unifying retrieve and rank with generative recommender and
  iterative preference alignment.
\newblock \emph{arXiv preprint arXiv:2502.18965}, 2025.

\bibitem[Feng et~al.(2021{\natexlab{a}})Feng, Gong, Sun, Ge, and
  Ou]{feng2021revisit}
Yufei Feng, Yu~Gong, Fei Sun, Junfeng Ge, and Wenwu Ou.
\newblock Revisit recommender system in the permutation prospective.
\newblock \emph{arXiv preprint arXiv:2102.12057}, 2021{\natexlab{a}}.

\bibitem[Feng et~al.(2021{\natexlab{b}})Feng, Hu, Gong, Sun, Liu, and
  Ou]{feng2021grn}
Yufei Feng, Binbin Hu, Yu~Gong, Fei Sun, Qingwen Liu, and Wenwu Ou.
\newblock Grn: Generative rerank network for context-wise recommendation.
\newblock \emph{arXiv preprint arXiv:2104.00860}, 2021{\natexlab{b}}.

\bibitem[Geng et~al.(2022)Geng, Liu, Fu, Ge, and Zhang]{geng2022p5}
Shijie Geng, Shuchang Liu, Zuohui Fu, Yingqiang Ge, and Yongfeng Zhang.
\newblock Recommendation as language processing (rlp): A unified pretrain,
  personalize, prompt and predict paradigm (p5).
\newblock In \emph{Proc.~RecSys'22}, 2022.

\bibitem[Gong et~al.(2021)Gong, Jiang, Wang, Lin, Feng, and
  Liang]{gong2021edge}
Yu~Gong, Xi~Jiang, Yuwei Wang, Lin Lin, Kai Feng, and Hongyan Liang.
\newblock Edge-cloud polarized reranking system for web-scale video
  recommendation.
\newblock In \emph{Proc.~SIGIR'21}, 2021.

\bibitem[Guo et~al.(2017)Guo, Tang, Ye, Li, and He]{guo2017deepfm}
Huifeng Guo, Ruiming Tang, Yunming Ye, Zhenguo Li, and Xiuqiang He.
\newblock Deepfm: a factorization-machine based neural network for ctr
  prediction.
\newblock In \emph{Proc.~IJCAI'17}, 2017.

\bibitem[Hidasi et~al.(2016)Hidasi, Karatzoglou, Baltrunas, and
  Tikk]{hidasi2016session}
Bal{\'a}zs Hidasi, Alexandros Karatzoglou, Linas Baltrunas, and Domonkos Tikk.
\newblock Session-based recommendations with recurrent neural networks.
\newblock In \emph{Proc.~ICLR'16}, 2016.

\bibitem[Ie et~al.(2019)Ie, Jain, Wang, Narvekar, Agarwal, Wu, Cheng, Chandra,
  and Boutilier]{ie2019slateq}
Eugene Ie, Vihan Jain, Jing Wang, Sanmit Narvekar, Ritesh Agarwal, Rui Wu,
  Heng-Tze Cheng, Tushar Chandra, and Craig Boutilier.
\newblock {SLATEQ}: A tractable decomposition for reinforcement learning with
  recommendation sets.
\newblock In \emph{Proc.~IJCAI'19}, 2019.

\bibitem[Jiang et~al.(2018)Jiang, Gowal, Mann, and Rezende]{jiang2018beyond}
Ray Jiang, Sven Gowal, Timothy~A Mann, and Danilo~J Rezende.
\newblock Beyond greedy ranking: Slate optimization via list-cvae.
\newblock \emph{arXiv preprint arXiv:1803.01682}, 2018.

\bibitem[Kang and McAuley(2018)]{kang2018self}
Wang-Cheng Kang and Julian McAuley.
\newblock Self-attentive sequential recommendation.
\newblock In \emph{Proc.~ICDM'18}, 2018.

\bibitem[Li et~al.(2018)Li, Shamaiah, Zhe, Liang, and Yang]{li2018fast}
Chengtao Li, Sriram Shamaiah, Zhen Zhe, Hongyan Liang, and Yiyang Yang.
\newblock Fast greedy map inference for determinantal point process to improve
  recommendation diversity.
\newblock In \emph{Proc.~NeurIPS'18}, 2018.

\bibitem[Li et~al.(2022)Li, Zhu, Liu, Su, Cai, Zhang, Tang, Xiao, and
  He]{li2022pear}
Yi~Li, Jieming Zhu, Weiwen Liu, Liangcai Su, Guohao Cai, Qi~Zhang, Ruiming
  Tang, Xi~Xiao, and Xiuqiang He.
\newblock {PEAR}: Personalized re-ranking with contextualized transformer for
  recommendation.
\newblock In \emph{Companion Proc.~TheWebConf'22}, 2022.

\bibitem[Lin et~al.(2025)Lin, Li, Dai, Bao, Lin, Yu, Zhang, and
  Zhao]{lin2025gref}
Zhijie Lin, Zhuofeng Li, Chenglei Dai, Wentian Bao, Shuai Lin, Enyun Yu,
  Haoxiang Zhang, and Liang Zhao.
\newblock {GReF}: A unified generative framework for efficient reranking via
  ordered multi-token prediction.
\newblock In \emph{Proc.~CIKM'25}, 2025.

\bibitem[Liu et~al.(2023{\natexlab{a}})Liu, Yang, Qi, and
  Wang]{liu2023learning}
Hanyang Liu, Shuai Yang, Feng Qi, and Shuaiwen Wang.
\newblock Learning to rank normalized entropy curves with differentiable window
  transformation.
\newblock \emph{arXiv preprint arXiv:2301.10443}, 2023{\natexlab{a}}.

\bibitem[Liu et~al.(2021)Liu, Sun, Ge, Pei, and Zhang]{liu2021variation}
Shuchang Liu, Fei Sun, Yingqiang Ge, Changhua Pei, and Yongfeng Zhang.
\newblock Variation control and evaluation for generative slate
  recommendations.
\newblock In \emph{Proc.~TheWebConf'21}, 2021.

\bibitem[Liu et~al.(2023{\natexlab{b}})Liu, Cai, He, Sun, McAuley, Zheng,
  Jiang, and Gai]{liu2023generative}
Shuchang Liu, Qingpeng Cai, Zhankui He, Bowen Sun, Julian McAuley, Dong Zheng,
  Peng Jiang, and Kun Gai.
\newblock Generative flow network for listwise recommendation.
\newblock In \emph{Proc.~KDD'23}, 2023{\natexlab{b}}.

\bibitem[Ma et~al.(2018)Ma, Zhao, Yi, Chen, Hong, and Chi]{ma2018modeling}
Jiaqi Ma, Zhe Zhao, Xinyang Yi, Jilin Chen, Lichan Hong, and Ed~H Chi.
\newblock Modeling task relationships in multi-task learning with multi-gate
  mixture-of-experts.
\newblock In \emph{Proc.~KDD'18}, 2018.

\bibitem[Meng et~al.(2019)Meng, Zhang, Xuan, Zhan, and
  Yang]{meng2019tensorized}
Shuchang Meng, Xiaolin Zhang, Minh-Thang Xuan, Eric Zhan, and Yiyang Yang.
\newblock Tensorized determinantal point processes for recommendation.
\newblock In \emph{Proceedings of the 25th ACM SIGKDD International Conference
  on Knowledge Discovery \& Data Mining (KDD)}, pages 1885--1894, 2019.

\bibitem[Ning and Slim()]{ning2011sparse}
X~Ning and G~Karypis Slim.
\newblock Sparse linear methods for top-n recommender systems.
\newblock In \emph{Proceedings of the 2011 IEEE 11th International Conference
  on Data Mining}, pages 497--506.

\bibitem[Pan et~al.(2020)Pan, Qian, Chen, Liang, and Yang]{pan2020purs}
Yushun Pan, Fu-Lai Qian, Li~Chen, Hongyan Liang, and Yiyang Yang.
\newblock Purs: Personalized unexpected recommender system for improving user
  satisfaction.
\newblock In \emph{Proceedings of the 14th ACM Conference on Recommender
  Systems (RecSys)}, pages 283--292, 2020.

\bibitem[Pang et~al.(2020)Pang, Xu, Ai, Lan, Cheng, and Wen]{pang2020setrank}
Liang Pang, Jun Xu, Qingyao Ai, Yanyan Lan, Xueqi Cheng, and Jirong Wen.
\newblock {SetRank}: Learning a permutation-invariant ranking model for
  information retrieval.
\newblock In \emph{Proceedings of the 43rd International ACM SIGIR Conference
  on Research and Development in Information Retrieval}, pages 499--508, 2020.

\bibitem[Pang et~al.(2025)Pang, Liu, Li, Zhu, Luo, Yu, Wu, Shen, Xia, Liu,
  Chen, and Wang]{pang2025higr}
Yunsheng Pang, Zijian Liu, Yudong Li, Shaojie Zhu, Zijian Luo, Chenyun Yu,
  Sikai Wu, Shichen Shen, Congying Xia, Yanchi Liu, Haifeng Chen, and Liang
  Wang.
\newblock {HiGR}: Efficient generative slate recommendation via hierarchical
  planning and multi-objective preference alignment.
\newblock \emph{arXiv preprint arXiv:2512.24787}, 2025.

\bibitem[Pei et~al.(2019)Pei, Zhang, Zhang, Sun, Lin, Sun, Wu, Jiang, Ge, Ou,
  et~al.]{pei2019personalized}
Changhua Pei, Yi~Zhang, Yongfeng Zhang, Fei Sun, Xiao Lin, Hanxiao Sun, Jian
  Wu, Peng Jiang, Junfeng Ge, Wenwu Ou, et~al.
\newblock Personalized re-ranking for recommendation.
\newblock In \emph{Proceedings of the 13th ACM conference on recommender
  systems}, pages 3--11, 2019.

\bibitem[Pi et~al.(2019)Pi, Bian, Zhou, Zhu, and Gai]{pi2019practice}
Qi~Pi, Weijie Bian, Guorui Zhou, Xiaoqiang Zhu, and Kun Gai.
\newblock Practice on long sequential user behavior modeling for click-through
  rate prediction.
\newblock In \emph{Proc.~KDD'19}, 2019.

\bibitem[Rajput et~al.(2023)Rajput, Mehta, Singh, Keshavan, Vu, Heidt, Hong,
  Tay, Tran, Samost, Kula, Chi, and Sathiamoorthy]{tiger}
Shashank Rajput, Nikhil Mehta, Anima Singh, Raghunandan Keshavan, Trung Vu,
  Lukasz Heidt, Lichan Hong, Yi~Tay, Vinh~Q. Tran, Jonah Samost, Maciej Kula,
  Ed~H. Chi, and Maheswaran Sathiamoorthy.
\newblock Recommender systems with generative retrieval.
\newblock In \emph{Proc.~NeurIPS'23}, 2023.

\bibitem[Ren et~al.(2024)Ren, Yang, Wu, Xu, Wang, and
  Zhang]{ren2024nonautoregressive}
Yuxin Ren, Qiya Yang, Yichun Wu, Wei Xu, Yalong Wang, and Zhiqiang Zhang.
\newblock Non-autoregressive generative models for reranking recommendation.
\newblock In \emph{Proc.~KDD'24}, 2024.

\bibitem[Rendle(2010)]{rendle2010factorization}
Steffen Rendle.
\newblock Factorization machines.
\newblock In \emph{Proc.~ICDM'10}, 2010.

\bibitem[Sarwar et~al.(2001)Sarwar, Karypis, Konstan, and
  Riedl]{sarwar2001item}
Badrul Sarwar, George Karypis, Joseph Konstan, and John Riedl.
\newblock Item-based collaborative filtering recommendation algorithms.
\newblock In \emph{Proc.~WWW'01}, 2001.

\bibitem[Shi et~al.(2023)Shi, Yang, Wang, Wu, Guan, Liao, Wang, Wang, and
  Wang]{shen2023pier}
Xiaowen Shi, Fan Yang, Ze~Wang, Xiaoxu Wu, Muzhi Guan, Guogang Liao, Yongkang
  Wang, Xingxing Wang, and Dong Wang.
\newblock {PIER}: Permutation-level interest-based end-to-end re-ranking
  framework in e-commerce.
\newblock In \emph{Proc.~KDD'23}, 2023.

\bibitem[Sun et~al.(2019)Sun, Liu, Wu, Pei, Lin, Ou, and
  Jiang]{sun2019bert4rec}
Fei Sun, Jun Liu, Jian Wu, Changhua Pei, Xiao Lin, Wenwu Ou, and Peng Jiang.
\newblock {BERT4Rec}: Sequential recommendation with bidirectional encoder
  representations from transformers.
\newblock In \emph{Proc.~CIKM'19}, 2019.

\bibitem[Tang and Wang(2018)]{tang2018personalized}
Jiaxi Tang and Ke~Wang.
\newblock Personalized top-n sequential recommendation via convolutional
  sequence embedding.
\newblock In \emph{Proc.~WSDM'18}, 2018.

\bibitem[Tay et~al.(2022)Tay, Tran, Dehghani, Ni, Bahri, Mehta, Qin, Hui, Zhao,
  Gupta, Schuster, Cohen, and Metzler]{tay2022dsi}
Yi~Tay, Vinh~Q. Tran, Mostafa Dehghani, Jianmo Ni, Dara Bahri, Sanket Mehta,
  Zhen Qin, Kai Hui, Zhe Zhao, Jai Gupta, Tal Schuster, William~W. Cohen, and
  Donald Metzler.
\newblock Transformer memory as a differentiable search index.
\newblock In \emph{Proc.~NeurIPS'22}, 2022.

\bibitem[Tomasi et~al.(2025)Tomasi, Fabbri, Lalmas, and
  Dai]{tomasi2025diffslate}
Federico Tomasi, Francesco Fabbri, Mounia Lalmas, and Zhenwen Dai.
\newblock Prompt-to-slate: Diffusion models for prompt-conditioned slate
  generation.
\newblock In \emph{Proc.~RecSys'25}, 2025.

\bibitem[van~den Oord et~al.(2017)van~den Oord, Vinyals, and
  Kavukcuoglu]{vqvae}
Aaron van~den Oord, Oriol Vinyals, and Koray Kavukcuoglu.
\newblock Neural discrete representation learning.
\newblock In \emph{Proc.~NeurIPS'17}, 2017.

\bibitem[Wang et~al.(2021{\natexlab{a}})Wang, Sun, Zhang, Tao, Ju, Gao, and
  Zhang]{wang2021sliding}
Bo~Wang, Fan Sun, Erli Zhang, Zhe Tao, Linju Ju, Yihang Gao, and Xiao-Yong
  Zhang.
\newblock Sliding spectrum decomposition for diversified recommendation.
\newblock In \emph{Proc.~KDD'21}, pages 3696--3704, 2021{\natexlab{a}}.

\bibitem[Wang et~al.(2019)Wang, Fang, Liu, Chen, Tao, Peng, Jin, and
  Tian]{wang2019sequential}
Fan Wang, Xiaomin Fang, Lihang Liu, Yaxue Chen, Jiucheng Tao, Zhiming Peng,
  Cihang Jin, and Hao Tian.
\newblock Sequential evaluation and generation framework for combinatorial
  recommender system.
\newblock \emph{arXiv preprint arXiv:1902.00245}, 2019.

\bibitem[Wang et~al.(2017)Wang, Fu, Fu, and Wang]{wang2017deep}
Ruoxi Wang, Bin Fu, Gang Fu, and Mingliang Wang.
\newblock Deep \& cross network for ad click predictions.
\newblock In \emph{Proc.~ADKDD'17}. 2017.

\bibitem[Wang et~al.(2021{\natexlab{b}})Wang, Shivanna, Cheng, Jain, Lin, Hong,
  and Chi]{wang2021dcn}
Ruoxi Wang, Rakesh Shivanna, Derek Cheng, Sagar Jain, Dong Lin, Lichan Hong,
  and Ed~Chi.
\newblock Dcn v2: Improved deep \& cross network and practical lessons for
  web-scale learning to rank systems.
\newblock In \emph{Proc.~TheWebConf'21}, 2021{\natexlab{b}}.

\bibitem[Wu et~al.(2024)Wu, Zheng, Qiu, Wang, Gu, Shen, Qin, Zhu, Zhu, Liu,
  Xiong, and Chen]{wu2024survey}
Likang Wu, Zhi Zheng, Zhaopeng Qiu, Hao Wang, Hongchao Gu, Tingjia Shen, Chuan
  Qin, Chen Zhu, Hengshu Zhu, Qi~Liu, Hui Xiong, and Enhong Chen.
\newblock A survey on large language models for recommendation.
\newblock volume~27, 2024.

\bibitem[Xi et~al.(2022)Xi, Liu, Zhu, Zhao, Dai, Tang, Zhang, Zhang, and
  Yu]{xi2022mir}
Yunjia Xi, Weiwen Liu, Jieming Zhu, Xilong Zhao, Xinyi Dai, Ruiming Tang,
  Weinan Zhang, Rui Zhang, and Yong Yu.
\newblock Multi-level interaction reranking with user behavior history.
\newblock In \emph{Proc.~SIGIR'22}, 2022.

\bibitem[Xia et~al.(2023)Xia, Eksombatchai, Pancha, Badani, Wang, Gu, Joshi,
  Farahpour, Zhang, and Zhai]{xia2023transact}
Xue Xia, Pong Eksombatchai, Nikil Pancha, Dhruvil~Deven Badani, Po-Wei Wang,
  Neng Gu, Saurabh~Vishwas Joshi, Nazanin Farahpour, Zhiyuan Zhang, and Andrew
  Zhai.
\newblock {TransAct}: Transformer-based realtime user action model for
  recommendation at {Pinterest}.
\newblock In \emph{Proc.~KDD'23}, 2023.

\bibitem[Zhai et~al.(2024)Zhai, Liao, Liu, Wang, Li, Cao, Gao, Gong, Gu, He,
  Lu, and Shi]{hstu}
Jiaqi Zhai, Lucy Liao, Xing Liu, Yueming Wang, Rui Li, Xuan Cao, Leon Gao,
  Zhaojie Gong, Fangda Gu, Jiayuan He, Yinghai Lu, and Yu~Shi.
\newblock Actions speak louder than words: trillion-parameter sequential
  transducers for generative recommendations.
\newblock In \emph{Proc.~ICML'24}, 2024.

\bibitem[Zhou et~al.(2018)Zhou, Zhu, Song, Fan, Zhu, Ma, Yan, Jin, Li, and
  Gai]{zhou2018deep}
Guorui Zhou, Xiaoqiang Zhu, Chenru Song, Ying Fan, Han Zhu, Xiao Ma, Yanghui
  Yan, Junqi Jin, Han Li, and Kun Gai.
\newblock Deep interest network for click-through rate prediction.
\newblock In \emph{Proc.~KDD'18}, 2018.

\end{thebibliography}
